\documentclass[10pt,twocolumn,letterpaper]{article}

\usepackage[pagenumbers]{cvpr} % To force page numbers, e.g. for an arXiv version

\newcommand{\xmark}{\ding{55}} % 叉号

\usepackage[accsupp]{axessibility}  % Improves PDF readability for those with disabilities.

\usepackage{pifont}
\usepackage{graphicx}
\usepackage{stfloats}
\usepackage{multirow}
\usepackage[table]{xcolor}

\definecolor{cvprblue}{rgb}{0.21,0.49,0.74}
\usepackage[pagebackref,breaklinks,colorlinks,allcolors=cvprblue]{hyperref}

\def\paperID{846} % *** Enter the Paper ID here
\def\confName{CVPR}
\def\confYear{2026}

\title{Garments2Look: A Multi-Reference Dataset for High-Fidelity \\ Outfit-Level Virtual Try-On with Clothing and Accessories}

\author{
    Junyao Hu\textsuperscript{1} \quad
    Zhongwei Cheng\textsuperscript{2} \quad
    Waikeung Wong\textsuperscript{1} \quad
    Xingxing Zou\textsuperscript{1 $^*$} \vspace{0.5em} \\
    \textsuperscript{1}The Hong Kong Polytechnic University \quad 
    \textsuperscript{2}Huhu AI Inc. \\ 
    \textsuperscript{$^*$}Corresponding author \\
    {\tt\small \{junyao.hu,calvin.wong,xingxing.zou\}@polyu.edu.hk, zcheng@huhu.ai}
}

\begin{document}

\maketitle

{
\lefthyphenmin=100
\abstract{Virtual try-on (VTON) has advanced single-garment visualization, yet real-world fashion centers on full outfits with multiple garments, accessories, fine-grained categories, layering, and diverse styling, remaining beyond current VTON systems. Existing datasets are category-limited and lack outfit diversity. We introduce Garments2Look, the first large-scale multimodal dataset for outfit-level VTON, comprising 80K many-garments-to-one-look pairs across 40 major categories and 300+ fine-grained subcategories. Each pair includes an outfit with 3-12 reference garment images (Average 4.48), a model image wearing the outfit, and detailed item and try-on textual annotations.
To balance authenticity and diversity, we propose a synthesis pipeline. It involves heuristically constructing outfit lists before generating try-on results, with the entire process subjected to strict automated filtering and human validation to ensure data quality.
To probe task difficulty, we adapt SOTA VTON methods and general-purpose image editing models to establish baselines. Results show current methods struggle to try on complete outfits seamlessly and to infer correct layering and styling, leading to misalignment and artifacts. Our code and data are open-sourced on~\url{https://github.com/ArtmeScienceLab/Garments2Look}.
}}
\vspace{-4mm}
    
\section{Introduction}
Virtual Try-On (VTON) has demonstrated significant application potential in fields such as e-commerce~\cite{sekri2025effects, gabriel2023influence}, visual effects~\cite{bug2019future}, fashion design~\cite{GenAI_in_fashion_Shi,lage2019virtual}, and human-computer interaction~\cite{prakash2024gesture}. 
At present, user's expectations for VTON are no longer limited to a single garment, but extend to the ability to intuitively and accurately preview more complex outfits.
Some research efforts have begun to investigate multiple items~\cite{Zhu_2024_CVPR_mmvto, Choi_2025_CVPR, chong2025fastfitacceleratingmultireferencevirtual, li2024anyfit}, layered garments~\cite{cui2021dressing, velioglu2025mgt, shao2023towards, yu2026go}, fine-grained categories~\cite{wan2025vtonvllm, Cui_2025_WACV}, and styling techniques~\cite{Zhu_2024_CVPR_mmvto, li2024controlling, kim2025promptdresser}, but no single method has emerged that focuses all these issues comprehensively.

\begin{figure}[t]
    \centering
    \includegraphics[width=1\linewidth]{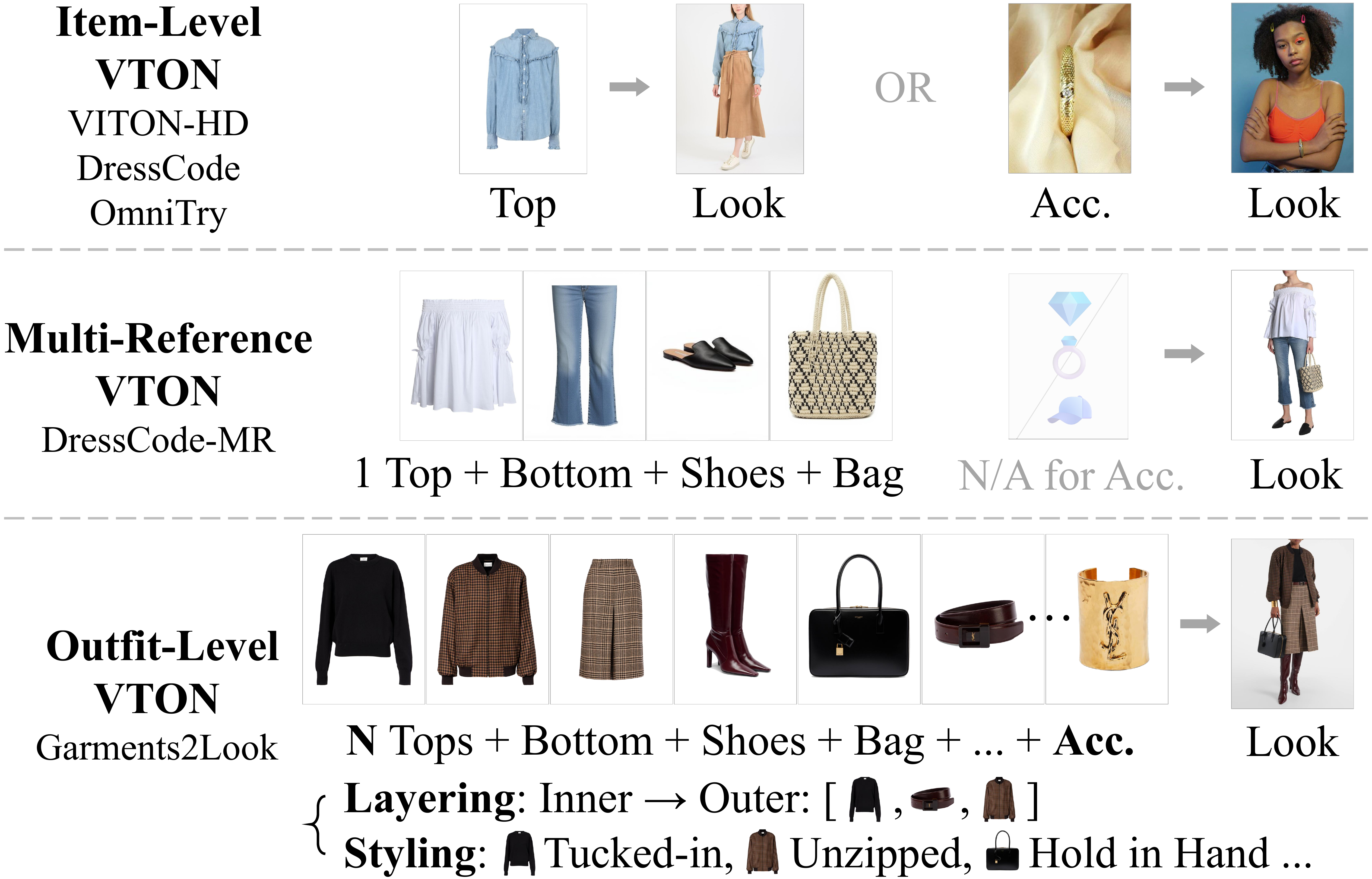}
    \vspace{-7mm}
    \caption{Comparison of data formats in virtual try-on datasets. Outfit-level dataset is collected and generated from a large-scale set of real images, each paired with diverse clothing and accessories, and including the information of outfit layering and styling.}
    \label{fig:data-format}
    \vspace{-1.9em}
\end{figure}

A direct reason for this limitation lies in the structural deficiencies of existing image VTON datasets. 
As shown in~\cref{fig:data-format}, representative datasets such as VITON-HD~\cite{choi2021viton} and DressCode~\cite{morelli2022dresscode} have advanced in image quality and scale but were originally designed solely for single-garment try-on tasks. 
They overlook the role of the accessory, and lack textual annotations such as dressing techniques (\eg, whether a shirt is tucked into pants) and inter-garment coordination relationships (\eg, layering order in an outfit).
Although OmniTry~\cite{feng2025omnitry} enriches the range of wearable categories, its task remains limited to individual items, while M\&M VTO~\cite{Zhu_2024_CVPR_mmvto}, BootComp~\cite{Choi_2025_CVPR} and DressCode-MR~\cite{chong2025fastfitacceleratingmultireferencevirtual} support multi-reference inputs but suffer from limited garment category diversity. 
Therefore, there is a clear need for a new virtual try-on dataset that simultaneously supports diverse item categories and coherent outfit-level composition.

Compared to single-item VTON, outfit-level VTON introduces new technical challenges.
Garments exhibit complex layering and occlusion relationships. For instance, inner-outer ordering varies (a thin knit cardigan may be the outermost layer or worn under a coat), and dressing hacks differ (it can be worn normally, draped over the shoulders, or cinched around the waist).
Faithfully capturing such details is vital to the quality and practical utility of the result.

To this end, we propose Garments2Look, paving the way for more advanced VTON that meets real-world needs. Our contributions are as follows:

\begin{itemize}
    \item We introduce a large-scale, multimodal, open-source dataset tailored for outfit-level VTON, covering a wide range of fashion items and comprising 80K high-quality item-model image pairs.
    \item We define a new VTON task that leverages rich structured annotations (text descriptions, layering order, dressing techniques) to apply multiple reference items along with their matching relationships to a model, producing flexible and diverse outfit try-on results.
    \item We conduct extensive experiments with state-of-the-art VTON methods and make an in-depth analysis to reveal their shortcomings and offer insights for improvement.
\end{itemize}
\begin{table*}[t]
\centering
\caption{Comparisons of image VTON datasets. We focus on outfit-level VTON. Our data is built upon real high-solution real-world garment and model images. Look Resolution = The resolution (height$\times$width) of target image. $\le$/$\sim$ = The number of pixel  is less than / approximately equal to the product. R/S = Real/Synthetic data. Layering/Styling = The dataset contains annotation of the garment layering order / styling techniques. VLM = Annotations are generated by visual-language models. Publicity = The dataset is open-sourced.}
\vspace{-0.5em}
\resizebox{0.99\textwidth}{!}{
\begin{tabular}{lllllllllcccc}
\toprule
Dataset Name &
  Time &
  Level &
  \#Class &
  \#Image &
  \#Item &
  \#Pair &
  \#Ref Image/Pair &
  Look Resolution &
  Source &
  Layering & 
  Styling & 
  Publicity \\
\midrule
VITON-HD~\cite{choi2021viton} & CVPR 21 & 
Item & 1 & 27358 & 13679 & 13679 & 1 & 
1024$\times$768 & R & \xmark & \xmark & \href{https://github.com/shadow2496/VITON-HD}{Link} \\
DressCode~\cite{morelli2022dresscode} & ECCV 22 & 
Item & 3 & 107584 & 53792 & 53792 & 1-2 (Avg 1.08) &
1024$\times$768 & R & \xmark & \xmark & \href{https://github.com/aimagelab/dress-code}{Link} \\
M\&M VTO~\cite{Zhu_2024_CVPR_mmvto} & CVPR 24 & 
Outfit & 2 & - & - & 18.8M & 2 &
1024$\times$512 & R & \xmark & 5 Types & \xmark \\
Street TryOn~\cite{Cui_2025_WACV} & WACV 25 &
Item & 13 & 14453 & 14453 & 0 & 1 &
512$\times$320 & R & \xmark & \xmark & \href{https://github.com/cuiaiyu/street-tryon-benchmark}{Link} \\ 
Shining Yourself~\cite{Miao_2025_CVPR} & CVPR 25 &
Item & 4 & - & - & 64000 & 1 &
512$\times$512 & R & \xmark & \xmark & \xmark \\
BootComp~\cite{Choi_2025_CVPR} & CVPR 25 &
Outfit & 7 & - & - & 54000 & 1-4 &
512$\times$384 & R+S & \xmark & \xmark & \xmark \\
Dresscode-MR~\cite{chong2025fastfitacceleratingmultireferencevirtual} & arXiv 2508 &
Outfit & 5 & 99260 & 71081 & 28179 & 1-4 (Avg 2.56) &
1024$\times$768 & R+S & \xmark & \xmark & \href{https://github.com/Zheng-Chong/FastFit}{Link} \\
Nano-Con OOTD~\cite{ye2025echo4o} & arXiv 2508 & 
Outfit & 6 & - & - & 19000 & 2-6 &
$\le$1024$\times$1024 & R+S & \xmark & VLM & \xmark \\
OmniTry-Train~\cite{feng2025omnitry} & arXiv 2508 &
Item & 12 & - & - & 51195 & 1 &
$\le$1024$\times$1024 & R+S & \xmark & VLM & \xmark \\
OmniTry-Bench~\cite{feng2025omnitry} & arXiv 2508 &
Item & 12 & 825 & 360 & 6975 & 1 &
$\sim$3000$\times$3000 & R & \xmark & VLM & \href{https://github.com/Kunbyte-AI/OmniTry}{Link} \\
GO-MLVTON~\cite{yu2026go} & ICASSP 26 &
Outfit & 2 & 10629 & 7091 & 3528 & 2 &
512$\times$384 & R & 2 Layers & \xmark & \href{https://upyuyang.github.io/go-mlvton/}{Link} \\
Garments2Look & CVPR 26 &
Outfit & 40 & 429054 & 184367 & 80041 & 3-12 (Avg 4.48) &
$\sim$1024$\times$1024 & R+S & 1-5 Layers & VLM & \href{https://github.com/ArtmeScienceLab/Garments2Look}{Link} \\
\bottomrule
\end{tabular}
}
\vspace{-1.2em}
\label{table:compare-dataset}
\end{table*}

\section{Related Work}

\subsection{Image-based Virtual Try-On Dataset}

As shown in~\cref{table:compare-dataset}, we review the mainstream and recently released datasets for the image-based virtual try-on task. 
VITON-HD~\cite{choi2021viton} significantly increase the resolution of virtual try-on images to a considerable size, but it only focuses a single gender (female) and a single clothing type (top). 
DressCode~\cite{morelli2022dresscode} and M\&M VTO~\cite{Zhu_2024_CVPR_mmvto} acknowledge the importance of full-body garments and extend the clothing types to three categories (top, bottom, and full). 
For accessory try-on, Shining Yourself~\cite{Miao_2025_CVPR} collects paired images covering four categories: bracelets, rings, earrings, and necklaces.
BootComp~\cite{Choi_2025_CVPR} proposes a try-off-based data synthesizing pipeline and a data filtering strategy.
DressCode-MR~\cite{chong2025fastfitacceleratingmultireferencevirtual} is built by CatVTON~\cite{chong2025catvton} and FLUX~\cite{labs2025flux1kontextflowmatching}, which considers five item categories, newly including shoes and bags. 
OmniTry~\cite{feng2025omnitry} further expanded the application scenarios of VTON by considering more wearable types, but its data remains single-item paired images. 
Nano-Consistent-150K~\cite{ye2025echo4o} includes a 19K VTON subset, but it ignores pose consistency.
GO-MLVTON~\cite{yu2026go} constructs a dataset to address a specific challenge of handling two upper garments.

To address the limitations of existing datasets, we propose a novel outfit-level dataset with multiple key advantages: it contains a large amount of high-quality real-world data, supports outfit-level reference input, provides results with 1M pixel-level resolution, includes textual annotations for item and outfit  descriptions, layering order, and styling techniques. Our dataset comprehensively surpasses previous state-of-the-art alternatives, and all data is publicly released to promote further research in VTON.

\subsection{Multi-Reference Image Data Synthesis}

Existing work on multi-reference image generation primarily targets different reference types: subjects, identities, styles, control signals, \etc. To construct paired data, these methods commonly rely on open-vocabulary models (\eg, Grounding DINO~\cite{liu2023grounding} and SAM2~\cite{ravi2024sam2}) to obtain layouts or segmentation results of subject instances as references~\cite{zhang2024mmtryon,chen2025xverse,mou2025dreamo}. Several data processing strategies have been proposed to avoid “copy-paste” artifacts. UNO~\cite{Wu_2025_ICCV} leverages a Subject-to-Image model to synthesize reference images. ComposeMe~\cite{qian2025composeme} builds a multi-image identity dataset that enables disentangled control over identity, hairstyle, and clothing (treated as a single holistic attribute rather than individual garments). USO~\cite{wu2025uso} and DreamOmni2~\cite{xia2025dreamomni2} generate style reference images and content reference images for the target image. MultiRef~\cite{chen2025multiref} generates different signals of an object via render engines. 
Recent works, like Pico-Banana-400K~\cite{qian2025picobanana400k}, MultiBanana~\cite{oshima2025multibanana}, MICo-150K~\cite{wei2025mico}, UniRef-Image-Edit~\cite{wei2026uniref} and FireRed-Image-Edit~\cite{team2026firered}, adopt advanced models such as Nano Banana series~\cite{comanici2025gemini}, Seedream series~\cite{seedream2025seedream} and Qwen Image Edit series~\cite{wu2025qwen} as core synthesis engines, collecting high-quality multi-reference images by carefully filtering.

Aligned with these concurrent studies, we leverage the paradigm of employing advanced editing models for data synthesis and filtering, a methodology that has emerged as a prevailing industry consensus for generating high-quality paired data. However, unlike general-purpose research, our work specializes in VTON, prioritizing full-outfit consistency and details like layering order and styling techniques that are often overlooked in general synthesis frameworks.

\section{Garments2Look Dataset}

\begin{figure*}[t]
    \centering
    \includegraphics[width=1.0\linewidth]{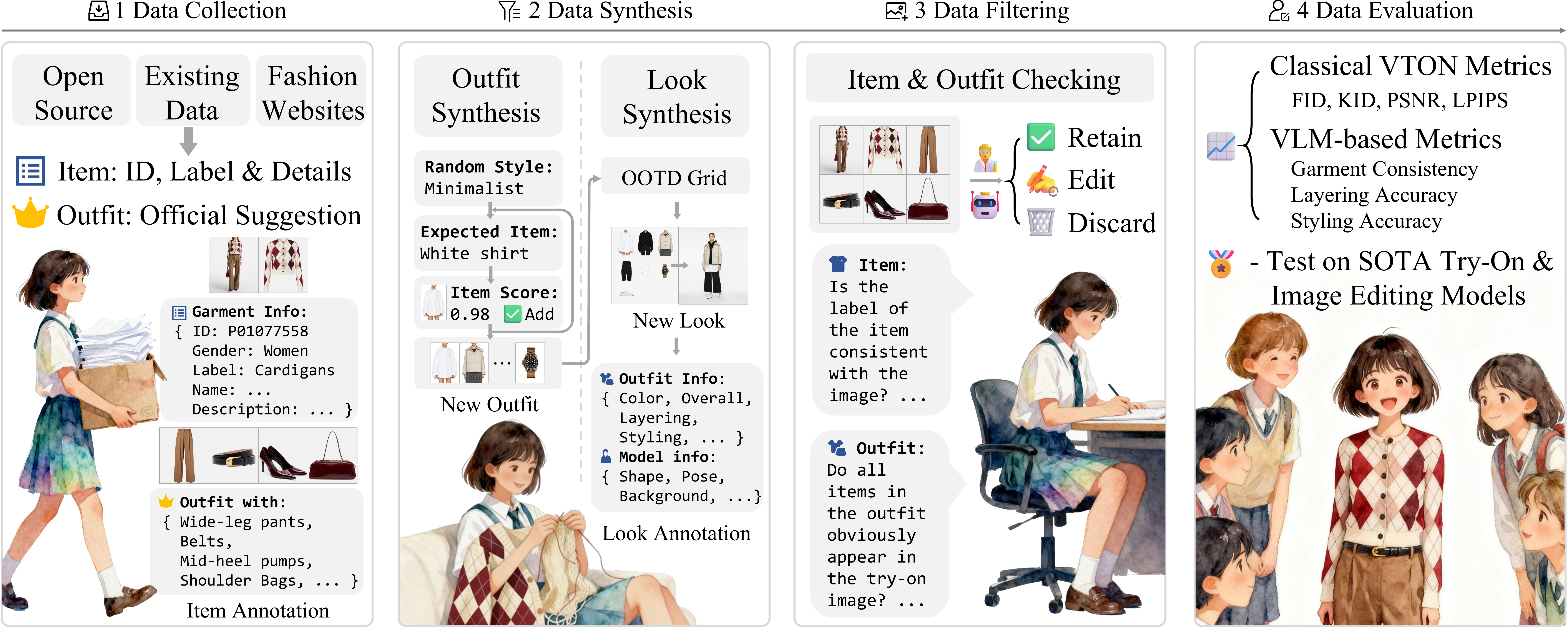}
    \vspace{-7mm}
    \caption{Overview of Garments2Look construction process.}
    \label{fig:overview}
    \vspace{-6mm}
\end{figure*}

As illustrated in~\cref{fig:overview}, the construction of Garments2Look follows four steps:
(1) Data Collection: obtaining real-world clothing items and their outfit suggestions from different sources;
(2) Data Synthesis: enriching the dataset content and diversity by generating new outfit lists and look images; 
(3) Data Filtering: ensuring visual consistency and data quality, including annotations of garment images, outfit lists and look images; 
and (4) Data Evaluation: verifying the data quality, designing new metrics for outfit-level VTON task, and testing SOTA models.

\subsection{Data Collection}
\label{sec:Collection}

To construct a dataset suitable for outfit-level VTON, we require paired input and output images:
the input consists of garment images for several individual items (such as multi-layered upper clothes, bottoms, accessories, \etc), and the output is a look image which makes the complete outfit coherently show on a human model.
However, perfectly matched paired data is often scarce and difficult to gather.
Therefore, we categorize the data based on its completeness and availability as follows:
(1) \textbf{Gold Standard Data}: Includes a set of garment images and their corresponding model-worn images, forming a naturally appropriate input-output pair. 
(2) \textbf{Garment Images with Paired Outfits}: Outfit composition without corresponding look image.
(3) \textbf{Garment Images without Paired Outfits}: Raw garment images without known information of outfit lists.
(4) \textbf{Only Look Images}: Only the look image is available, with no relevant reference garment image provided.
In order to balance the trade-off between data quality and quantity, we primarily integrates data in Category 1 (50.2\%), 2 (24.0\%), and 3 (25.8\%):
On the one hand, we leverage high-quality gold standard data to ensure try-on fidelity (the model needs to know what ``real'' looks like). 
On the other hand, for unpaired images, we employ them to enhance the amount and diversity of our dataset via a data synthesis pipeline (see~\cref{sec:Synthesis}).
Our data is mainly sourced from four complementary streams:
(1) Foundation works in outfit compatibility learning~\cite{kaicheng2021modeling, zou2022good}.
(2) Curated open-source fashion datasets, \eg, Maryland PolyVore~\cite{han2017learning}, which provides high-quality and trustworthy outfit data. 
(3) Publicly available web images, with rigorous compliance to licensing and privacy permissions.
(4) Synthetic data generated by image generation models and image understanding models.

\subsection{Data Synthesis}
\label{sec:Synthesis}

Our data synthesis primarily focuses on two aspects:
(1) \textbf{Outfit Synthesis}: To utilize unpaired garment images, we adopt an approach similar to retrieval-augmented generation to heuristically construct outfit data.
(2) \textbf{Look Synthesis}: To utilize both existing non-gold-standard outfit data and newly synthesized outfit data, we use image generation models to synthesize try-on look results, and use image understanding models to generate detailed annotations.

\subsubsection{Outfit Synthesis}

\textbf{Outfit Synthesis Pipeline Overview}: We first randomly select a style from the pre-constructed fashion style knowledge base to serve as the generation anchor. Subsequently, a large language model (LLM) generates a detailed description of a potential user scenario and preferences based on the chosen style. The LLM then uses the context, combined with the style knowledge, to generate an outfit list. For each item in the list, we perform image retrieval to identify the most relevant items in the database. A re-weighted sampling strategy is applied to finally select suitable items.

\textbf{Step 1 - Outfit Knowledge Base Construction}:
To ensure that the synthesized data encompasses a wide spectrum of fashion styles while maintaining clear boundaries between them, we adopted a strategy combining outfit style guidance generation and fashion expert review to build the outfit style knowledge base.
The knowledge base covers 65 prevalent and subcultural fashion styles (35 for women and girls, 30 for men and boys), like Y2K Style, Fresh Style, Preppy Style, \etc. 
For each style, we first instruct the LLM to strictly follow a predetermined outline and Markdown structure to generate a technical style guide. 
This guide meticulously defines the style's preferences, prohibitions, classic pairing examples, and extended styling rules. 
Subsequently, fashion experts review and refine this guide, resulting in precise style prompts and knowledge files that are ultimately used to constrain the generation model. 

\textbf{Step 2 - User-Driven Context Generation}:
User context serves as the driving force for outfit synthesis.
To ensure the generated outfits possess practical relevance and high diversity, based on the randomly selected style and user gender, we prompt the LLM to heuristically imagine and create a diverse user profile and specific dressing context. 
These attributes include user demographics (\eg, age, occupation, interests) and the precise occasion (\eg, evening gala, casual outing). 
The context description encompasses four key dimensions: occasion, palette, theme, and garment types, thereby guaranteeing the contextual appropriateness of the subsequent outfit list generation.

\textbf{Step 3 - Outfit List Generation}:
Upon obtaining the detailed context and style knowledge, we utilize the LLM for outfit list generation.
The model is explicitly constrained to strictly adhere to user requirements and the style guide, outputting a complete outfit list comprising 3 to 9 individual items.
To simulate the complexity of real-life fashion, for example, we specifically instruct the model to focus on layering, allowing for a maximum of three layered tops in a combination. 
The generated list should follow a top-down, inner-to-outer, and garment-to-accessory order, ensuring the logical coherence and hierarchical sense.

\textbf{Step 4 - Item Retrieval}:
For each item description within the LLM-generated outfit list, we query it to fetch the top 128 most relevant items of the corresponding category from the image database, forming the candidate set. 
To address the issue of certain items being overlooked due to platform data bias, we introduce a re-weighted sampling mechanism to improve traditional similarity-driven selection.
We adjust the sampling probability of retrieval candidates according to their historical selection frequency: 
An item's selection probability is inversely proportional to how many times it has appeared in outfit data.
Items with lower historical usage thus receive a correspondingly higher selection chance.
This strategy discourages repeated selection of popular items, ensuring a more uniform item distribution across the corpus and improving the utilization of raw data.

More details about style guidance and retrieval sampling strategy can be seen in~\cref{supp:sec:outfit-synthesis}.

\subsubsection{Look Synthesis}
\label{sec:look-synthesis}

To convert non-gold-standard outfit data and synthesized outfit data into look image, we generated it based on the outfit-of-the-day (OOTD) grid image. 
We arrange all item images in an outfit list into a two-dimensional matrix grid figure, which is used as input for image generation mainly by Nano Banana (Gemini-2.5-Flash-Image)~\cite{comanici2025gemini}.
Compared to directly using multiple images as input, OOTD image as input can maintain better consistency between items (See~\cref{sec:garment2look-challenging}). 
We further investigated the impact of item position variations, random arrangements, and arrangements based on prior positions within the OOTD image, but observed no significant impact on the quality of final look image. 
To enhance the creativity and visual appeal of look images, we explicitly incorporate layering order and styling techniques via prompt engineering.
For layering order, we specify the exact garment order.
We adopt five types of styling techniques from previous work~\cite{li2024controlling, kim2025promptdresser, Zhu_2024_CVPR_mmvto, chen2023size, tran2025dualfit}, \eg, ``tucking in the top'' and ``rolled up the sleeves''.
We either specify the desired layering order and styling techniques, or let the model apply appropriate ones freely.

Furthermore, by taking look images as input, VLM provides richer textual descriptions, yielding more information for the textual modality of our dataset (See~\cref{sec:supp:look-synthesis}).

\subsection{Data Filtering}
\label{sec:filtering}
To ensure data quality, with the help of experts in fashion, we conducted data screening across three aspects: individual item images, outfit lists, and garments-look pairs.

As for individual item images, based on the metadata and common types widely adopted in existing  works~\cite{Cui_2025_WACV, Miao_2025_CVPR, Choi_2025_CVPR, chong2025fastfitacceleratingmultireferencevirtual, feng2025omnitry}, we defined 40 primary clothing and accessories categories, comprising 300+ fine-grained subcategories. 

As for outfit lists, although certain raw data provides pre-defined outfit lists, and our outfit synthetics pipeline can generate lists, these may be with logical redundancy (\eg, it is uncommon for a person to wear two dresses simultaneously).
To address this, we designed a rule-based outfit plausibility validation mechanism grounded in fashion expertise. 
In cases where an outfit violates the constraints, we extract subsets by removing redundant items.

As for garments-look image pairs, we focused on identifying and retaining two types of images: 
full garment images which clearly display the entire garment, and look images which completely display the model wearing the entire outfit, captured from a frontal viewpoint.
We utilized Gemini-2.5-Flash~\cite{comanici2025gemini} to filter suitable images and we also use tools like DWPose~\cite{yang2023effective} to classify look images.

To guarantee the quality of the synthetic data, we recruited 10 fashion students and 3 experts in the process.
If any garment within an outfit is inconsistent, the look image is regenerated or discarded. 
Only $\sim$40\% of the synthetic look images were included in the final dataset, with every single image passing expert review.

More details about primary garment categories and fashion expert review process are in~\cref{supp:sec:filtering}.

\begin{figure*}[t]
  \centering
  \subfloat[Number of outfit list by data division.]
  { 
    \vspace{-0.1em} 
    \includegraphics[width=0.142\textwidth]{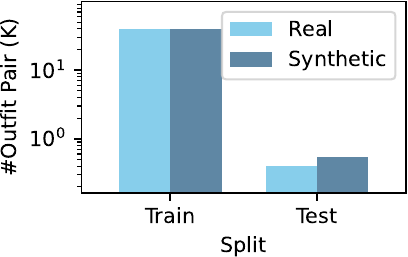}
    \label{fig:train_test_real_synthetic}
  }
  \hspace{0.1em}
  \subfloat[Number of outfit list by gender type.]
  {
    \includegraphics[width=0.142\textwidth]{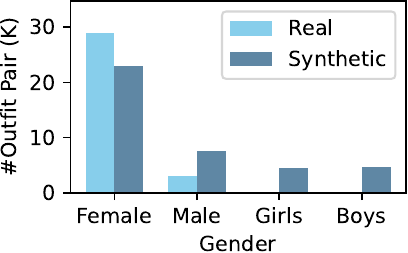}
    \label{fig:gender_distribution}
  }
  \hspace{0.1em}
  \subfloat[Number of outfit list by outfit length.]
  { 
    \includegraphics[width=0.142\textwidth]{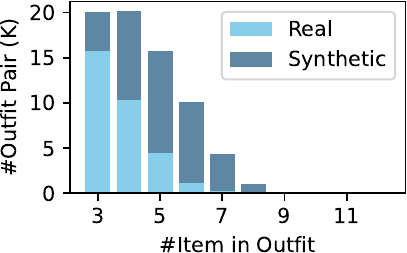}
    \label{fig:outfit_len_distribution}
  }
  \hspace{0.1em}
  \subfloat[Number of layering order by length.]
  {
    \includegraphics[width=0.142\textwidth]{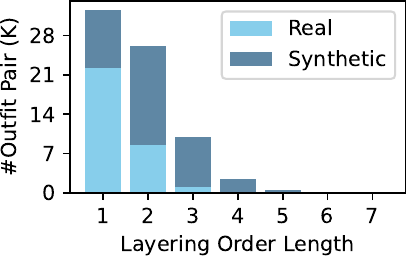}
    \label{fig:layering_order_length_distribution}
  }
  \hspace{0.1em}
  \subfloat[Top 10 clothing categories by item count, group by gender.]
  { 
    \vspace{0.35em}
    \includegraphics[width=0.3\textwidth]{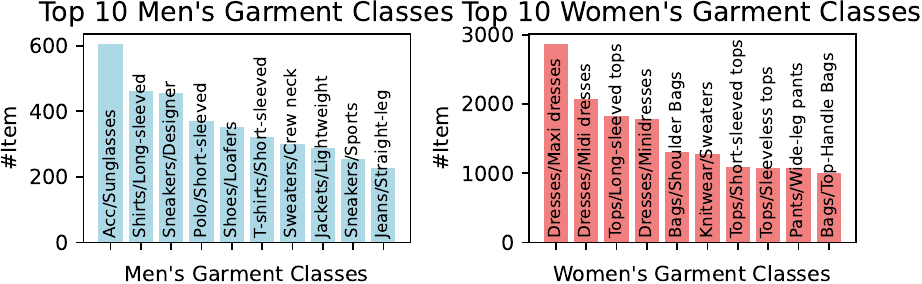}
    \label{fig:top10_garment_classes_by_gender}
  }
  \vspace{0.4em}
  \quad
  \subfloat[The number of top 8 types of outfit list.]
  { 
    \includegraphics[width=0.13\textwidth]{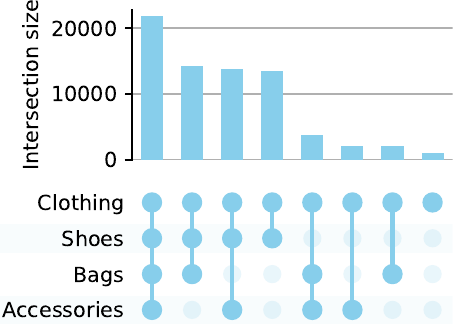}
    \label{fig:upset_outfit_type}
  }
  \hspace{0.4em}
  \subfloat[Word cloud of garment image description.]
  {
    \includegraphics[width=0.14\textwidth]{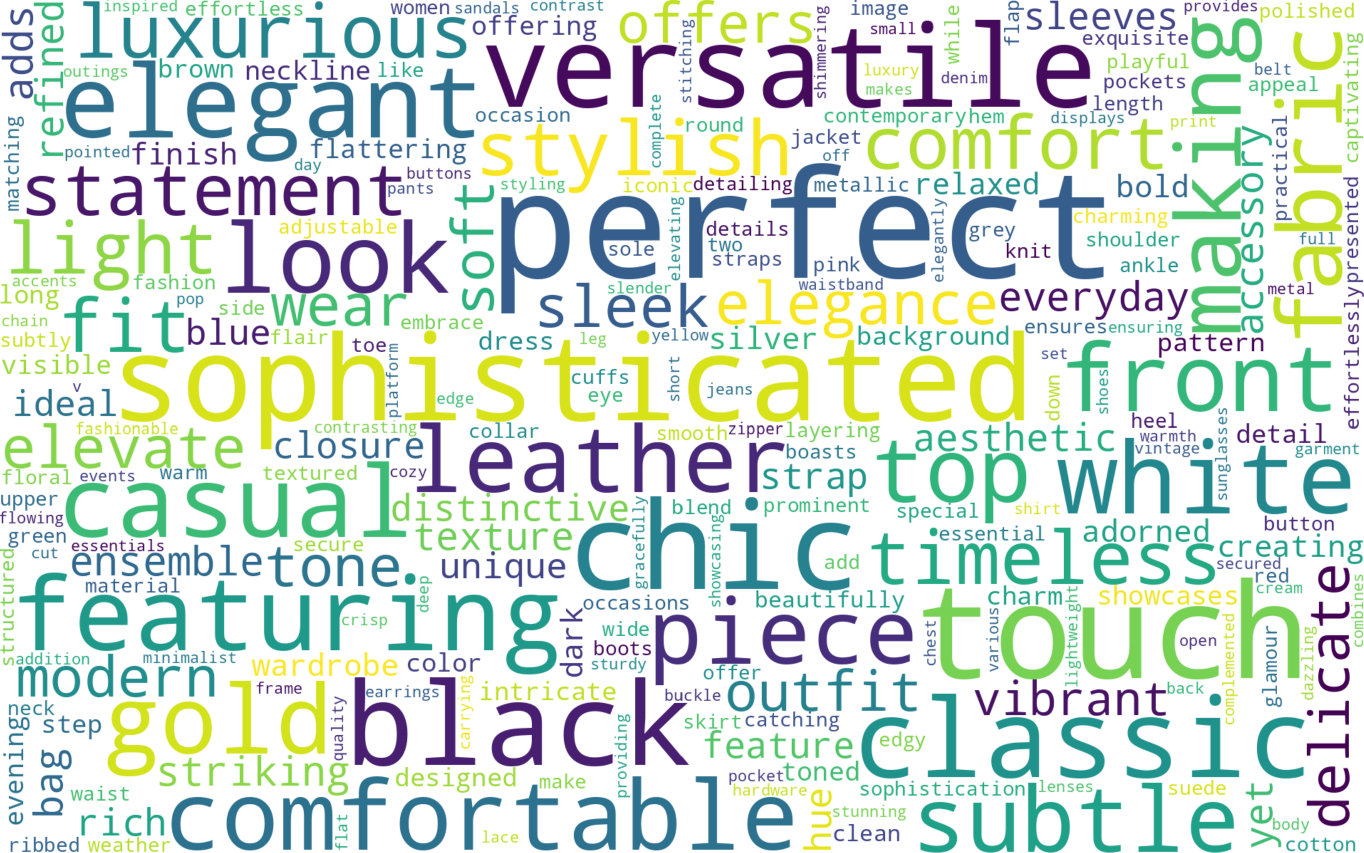}
    \label{fig:wordcloud_item_description}
  }
  \hspace{0.1em}
  \subfloat[Word cloud of look image description.]
  {
    \includegraphics[width=0.14\textwidth]{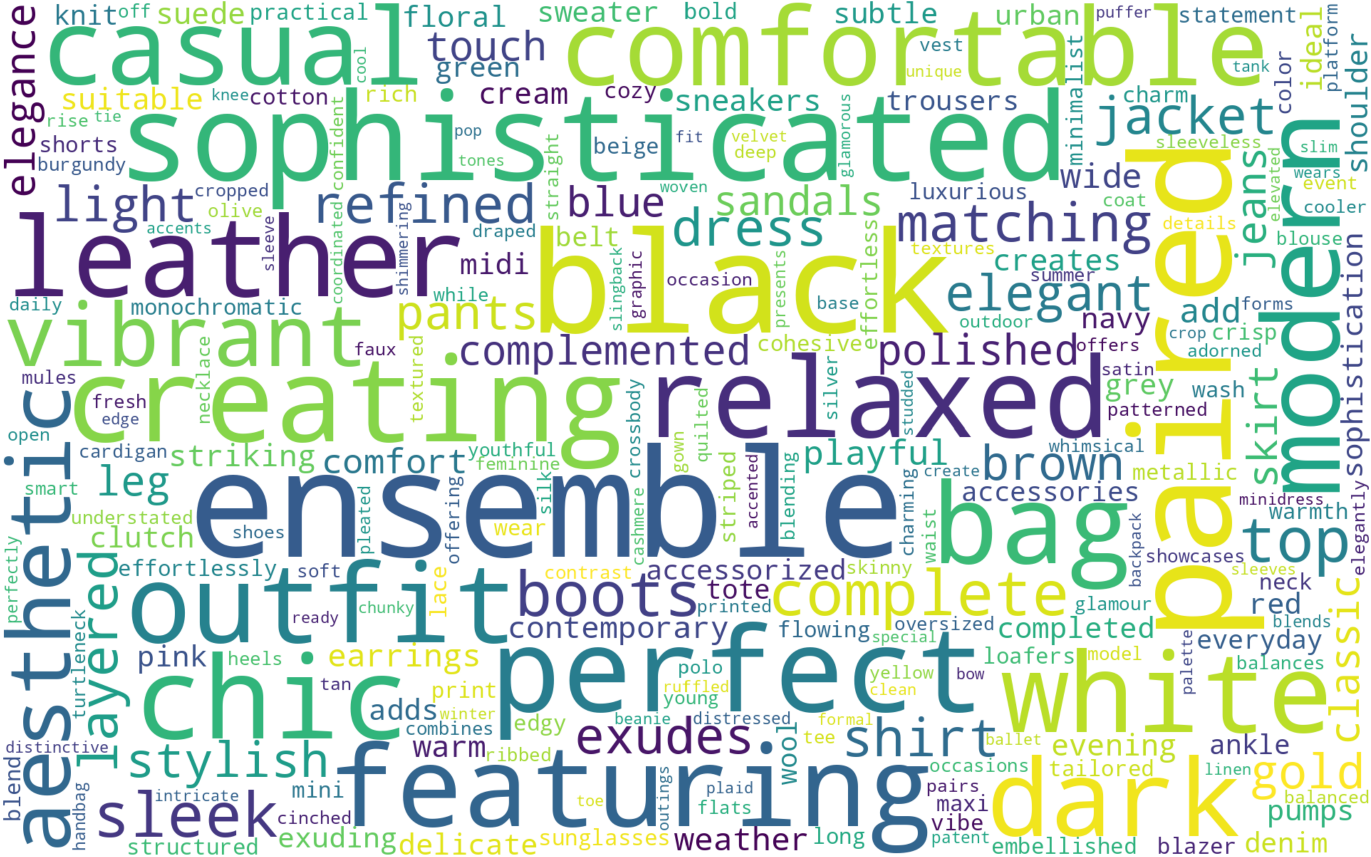}
    \label{fig:overall_look_description_wordcloud}
  }
  \hspace{0.1em}
  \subfloat[Word cloud of styling description.]
  {
    \includegraphics[width=0.14\textwidth]{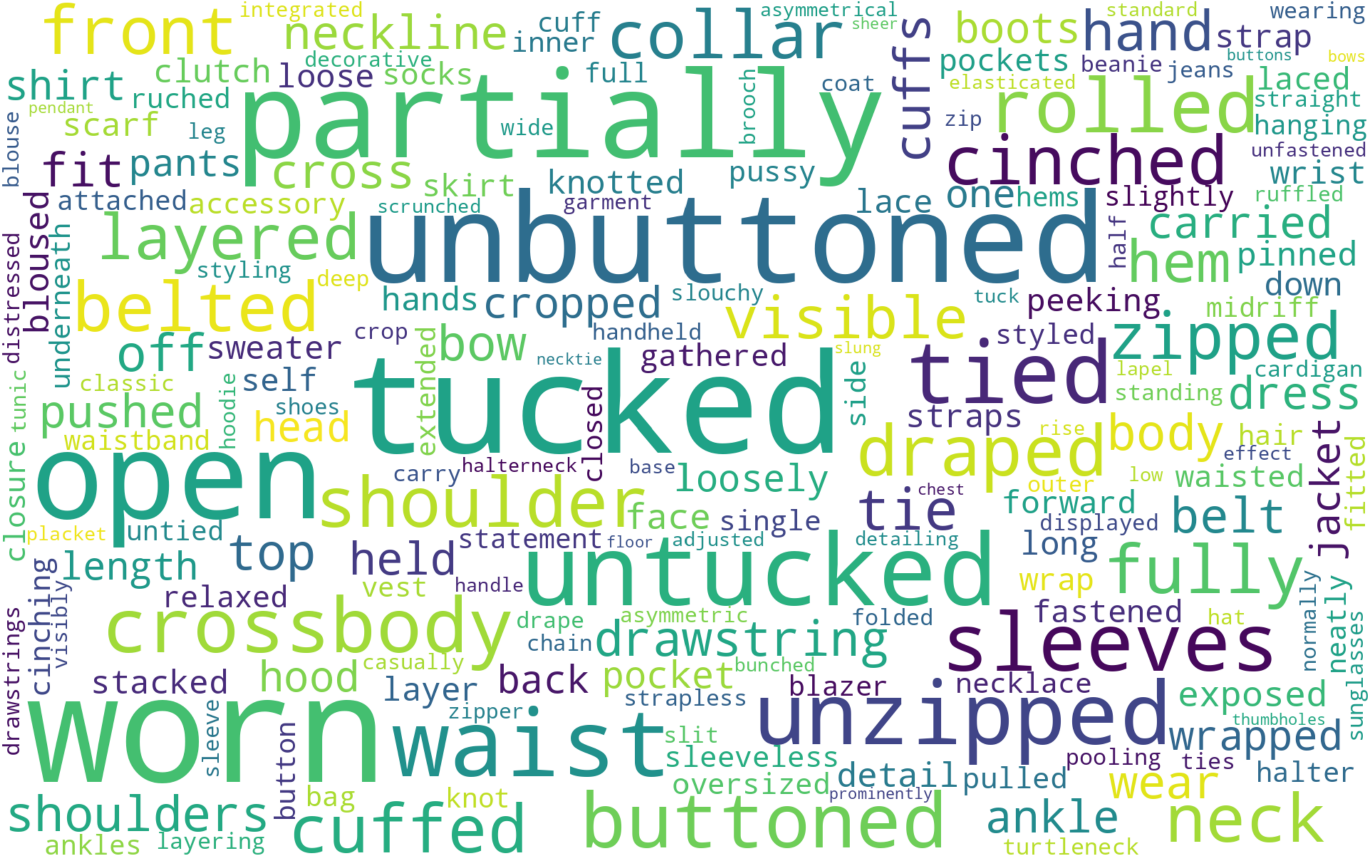}
    \label{fig:styling_techniques_wordcloud}
  }
  \hspace{0.4em}
  \subfloat[Aesthetic scores of sampled look images.]
  { 
    \includegraphics[width=0.14\textwidth]{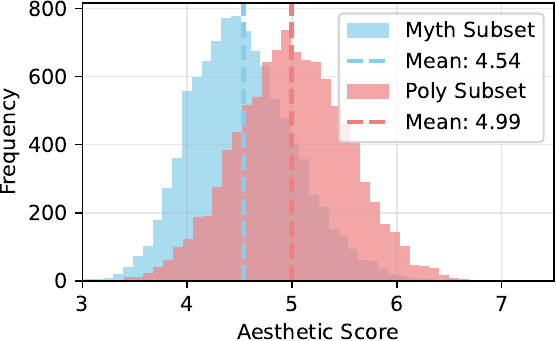}
    \label{fig:aesthetic-score}
  }
  \hspace{0.4em}
  \subfloat[Human evaluation on sampled outfit lists.]
  { 
    \vspace{0.6em}
    % \begin{table}[tbh]
\centering
\small
\setlength{\tabcolsep}{2.2pt}
\resizebox{0.12\textwidth}{!}{
\begin{tabular}{ccc}
\toprule
Eval Type & Mean $\pm$ SD \\
\midrule
Garments & 4.74 $\pm$ 0.28 \\
Layering  & 4.35 $\pm$ 0.62 \\
Styling  & 4.58 $\pm$ 0.45 \\
\bottomrule
\end{tabular}
}
% \vspace{-1.3em}
% \captionof{table}{\footnotesize Human evaluation.}
% \label{tab:human_eval}
% \end{table}
    \label{tab:human_eval}
  }
  \vspace{-0.8em}
  \caption{Data distribution statistics for Garments2Look.}
  \vspace{-1.8em}
  \label{fig:basic-info}
\end{figure*}

\subsection{Data Evaluation}
\label{sec:Evaluation}

% 数据占比、多样性
\textbf{Statistical analysis}: Garments2Look includes 80K outfit-level pairs, and \cref{fig:basic-info} presents basic statistics of the dataset.
The real and synthetic data in the final dataset are maintained at $\sim$1:1 ratio (\cref{fig:train_test_real_synthetic}). 
We collect data covering diverse genders (\cref{fig:gender_distribution}), different numbers of garment images per outfit (\cref{fig:outfit_len_distribution}), different layering order lengths (\cref{fig:layering_order_length_distribution}), and encompassing a broad range of garments categories (\cref{fig:top10_garment_classes_by_gender}) and outfit combination patterns (\cref{fig:upset_outfit_type}). 
% 
% 文本模态 多样性
We also pay attention to textual annotation to facilitate future multimodal research, including descriptions of item images, look images, and styling techniques.
In~\cref{fig:wordcloud_item_description,fig:overall_look_description_wordcloud,fig:styling_techniques_wordcloud}, these three word clouds illustrate the three core dimensions of text annotations within our dataset. Garment descriptions emphasize intrinsic attributes and textures; high-frequency terms such as ``leather'', ``elegant'', and ``sophisticated'' indicate that these annotations are designed to characterize material properties, styles, and design details. Look annotations focus on high-level visual effects and coordination. Keywords like ``ensemble'', ``relaxed'', and ``chic'' highlight the holistic look on the model. Furthermore, styling descriptions prioritize specific wearing states. The prevalence of action-oriented verbs, such as ``tucked'', ``unbuttoned'', and ``rolled'', clearly reflects a focus on the physical interaction between the garment and the body. Collectively, these multi-dimensional textual cues provide comprehensive guidance for achieving high-fidelity and precise VTON.
We also use aesthetic-predictor-v2-5~\cite{aesthetic_predictor_v2_5} to assess the aesthetic quality of look images.
While prior works~\cite{schuhmann2022laion} commonly adopt an absolute aesthetic threshold of 5.0, such a fixed cutoff may be suboptimal for human-centric fashion imagery. 
Hence, we filtered images with aesthetic scores below the empirical mean of each dataset subset, thereby removing clearly low-quality outputs. 
The remaining candidates are then subjected to manual filtering.
In~\cref{fig:aesthetic-score}, we finally evaluate 10K samples each from the 2 subsets of Garments2Look.
As for consistency and accuracy, in~\cref{tab:human_eval}, we ask 13 fashion experts to assess the consistency and accuracy of randomly selected 100 samples in the training set based on a Likert scale score (1-5). The higher the score, the greater the degree of consistency or accuracy.

\textbf{Outfit-level VTON Evaluation Protocol}:
For automatic evaluation on the performance of models, classical VTON metrics are considered: FID~\cite{parmar2022aliased}, KID~\cite{sutherland2018demystifying}, SSIM~\cite{sutherland2018demystifying}, and LPIPS~\cite{zhang2018unreasonable}. 
For our outfit-level VTON task, We leverage Gemini-3-Flash as a VLM judge to evaluate results across three metrics, reporting binary classification accuracy. Garment consistency is evaluated per item; partial visibility due to occlusion is accepted, while structural mismatches (\eg, wrong pocket geometry and position) are considered as inconsistent. Layering accuracy is optimized to linear complexity by verifying inner-outer relationships only between adjacent layers. Styling accuracy is similarly assessed on each garment. 
All the evaluation results need to simultaneously output both the classification results and the reasons to ensure interpretability. 

More details about the dataset are in~\cref{sec:supp:details}.

\section{Experiments}

We design experiments to validate the valuables of the proposed Garments2Look from two aspects:
(1) \textbf{Dataset difficulty}: existing models underperform on outfit-level VTON with accessories, layering orders and styling techniques. 
(2) \textbf{Actionable insights}: beyond-visual structured annotations (layering order, styling techniques, and more textual descriptions) can provide effective guidance to improve generation.
To this end, we first benchmark a strong image editing model on an established multi-reference VTON dataset to calibrate its upper bound in a simpler setting.
We then evaluate both VTON models and general-purpose editing models on test set of Garments2Look, qualitatively and quantitatively, to expose bottlenecks and analyze how our structured annotations can help.

\subsection{Can Editing Models Work on VTON?}
\label{sec:can-work}

\begin{table}[t]
\caption{Quantitative comparison on DressCode-MR~\cite{chong2025fastfitacceleratingmultireferencevirtual} for multi-reference VTON. We evaluate the performance of Nano Banana (Gemini-2.5-Flash-Image)~\cite{comanici2025gemini} by two reference strategies.}
\vspace{-0.5em}
\resizebox{0.48\textwidth}{!}{
\begin{tabular}{lcccccc}
\toprule
\multirow{2}{*}[-0.2em]{Method} & \multicolumn{4}{c}{Paired}     & \multicolumn{2}{c}{Unpaired} \\ \cmidrule(r){2-5} \cmidrule(l){6-7} 
& FID$\downarrow$ & KID$\downarrow$  & SSIM $\uparrow$ & LPIPS$\downarrow$ & FID$\downarrow$ & KID$\downarrow$ 
                        \\ \midrule
CatVTON~\cite{chong2025catvton} & 16.131 & 6.980 & \underline{0.856} & 0.106 & 18.339        & 7.458        \\
IP-Adapter~\cite{ye2023ip-adapter} & \underline{14.459} & \underline{4.144} & 0.861 & \underline{0.089} & 24.139        & 10.783       \\
FitDiT~\cite{jiang2024fitditadvancingauthenticgarment} & 14.722 & 5.471 & 0.850 & 0.122 & 15.956        & \underline{5.645}        \\
FastFit~\cite{chong2025fastfitacceleratingmultireferencevirtual} & \textbf{9.311}  & \textbf{1.512} & \textbf{0.859} & \textbf{0.079} & \textbf{12.059} & \textbf{2.123} \\
Nano-Banana~\cite{comanici2025gemini} (N Ref) & 15.220 &  6.703  & 0.801 & 0.199 & \underline{15.369} & 6.916 \\
Nano-Banana~\cite{comanici2025gemini} (2 Ref) & 15.980 & 7.494 &     0.797 & 0.231 &  16.393 & 7.535 \\
\bottomrule
\end{tabular}
}
\label{table:dresscode-mr-compare}
\vspace{-1.5em}
\end{table}

With the rapid progress of recent general-purpose editing models, their performance has attracted substantial attention. VTON as a task with high practical value, is frequently adopted by commercial vendors to showcase editing capabilities; VTON scenarios appear prominently in many promotional materials for these general-purpose editing products. This raises an apparently paradoxical question: if contemporary commercial editing models already handle multi-item try-on effectively, does our proposed dataset and task still matter? Conversely, if these commercial models cannot yet solve multi-item try-on, how can we conveniently and scalably construct large multi-item datasets? To address this, we design our first set of studies: can advanced image editing models synthesize usable multi-reference try-on data on existing datasets? The goal is to establish a feasibility baseline on a relatively simple multi-reference benchmark, clarify why current editing models remain insufficient for our more challenging setting, and highlight their promise for scalable data synthesis.

Concretely, we evaluate SOTA image editing model, Nano Banana (Gemini-2.5-Flash-Image)~\cite{comanici2025gemini}, on the latest dataset tailored for multi-reference VTON, DressCode-MR~\cite{chong2025fastfitacceleratingmultireferencevirtual}. As reported in~\cref{table:dresscode-mr-compare}, Nano Banana still lags behind the best VTON-specific methods. 
At first glance this is counterintuitive, since, as described in~\cref{sec:look-synthesis}, we employ Nano Banana for look image synthesis. 
Our model choice was supported by expert cross-validation: three fashion experts compared 500 results across two VTON models (OmniTry~\cite{feng2025omnitry} and FastFit~\cite{chong2025fastfitacceleratingmultireferencevirtual}) and three editing models (Nano Banana~\cite{comanici2025gemini}, Seedream 4.0~\cite{seedream2025seedream}, and Qwen-Image-Edit-2509~\cite{wu2025qwen}); Nano Banana was preferred in 66\% of cases. 
Upon closer analysis, we find two key limitations: Nano Banana does not natively support image inpainting, and it lacks accurate explicit control over skeletal pose. As a result, despite compelling overall visual quality, its quantitative metrics and detail fidelity fall short of VTON-specialized models due to limited fine-grained controllability. 
Even when we provide a skeleton image as one of the references, image editing models fail to strictly preserve the same pose (See~\cref{sec:pose-condition}), since standard VTON metrics are sensitive to pose consistency and pose preservation is a long-standing requirement in VTON.

In summary, the absence of image inpainting and explicit pose control causes Nano Banana to underperform specialized VTON systems in fidelity and structural consistency. In other words, for editing models to robustly solve VTON, they must improve generation consistency under a fixed target identity and conditioning. On the other hand, for synthetic data generation, directly editing the entire image via prompts, without relying on skeleton or inpainting masks, possible to yield high-quality multi-item results. Given the high cost of collecting high-quality multi-reference data and our expert vetting of the synthetic results, these factors highlight the challenges of constructing such data and further underscore the value of this work.

\subsection{How Challenging Garments2Look is?}
\label{sec:garment2look-challenging}

\begin{figure*}[t]
    \centering
    \includegraphics[width=1\linewidth]{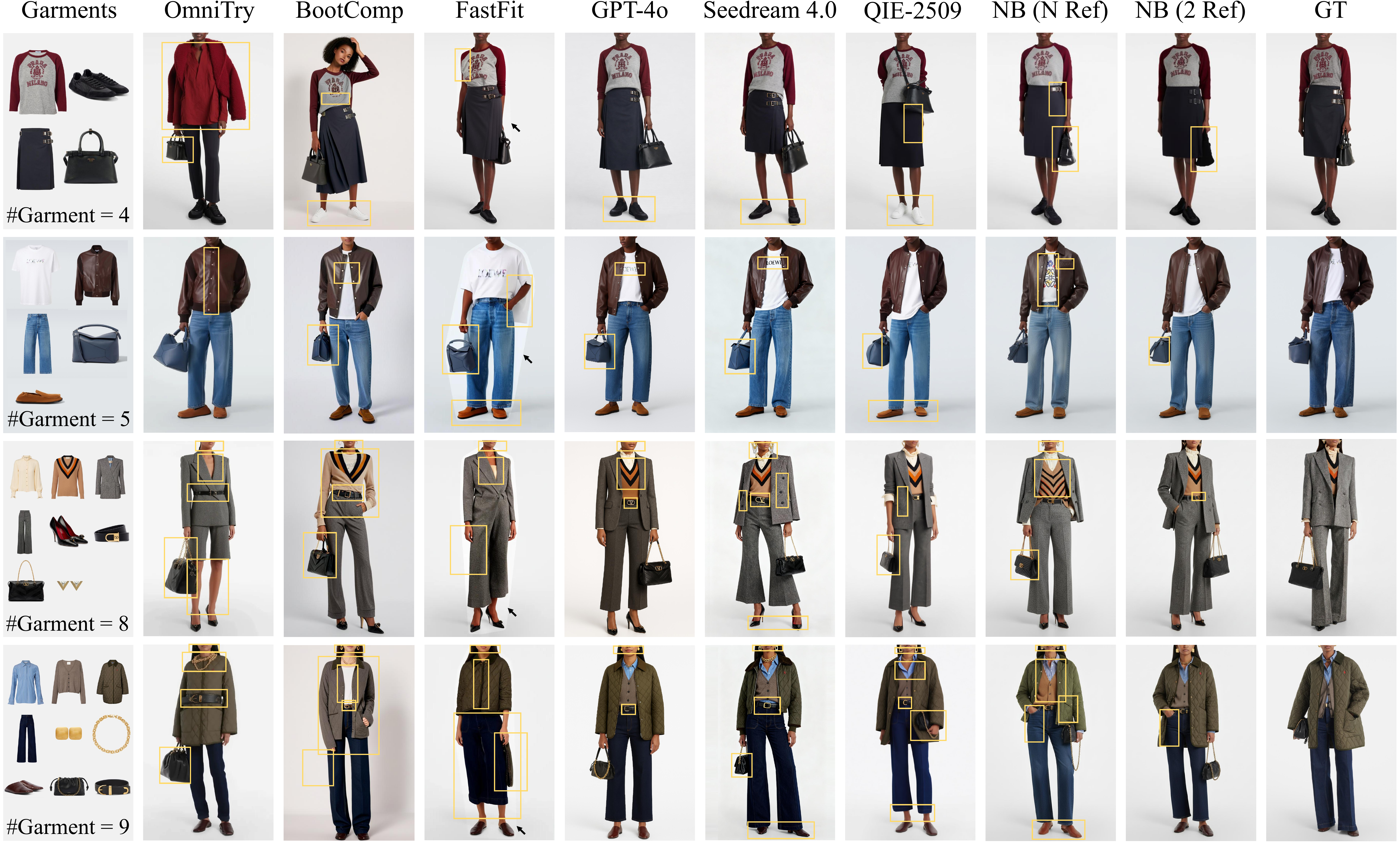}
    \vspace{-2em}
    \caption{Comparison of results of 3 SOTA VTON models and 4 general-purpose image editing models on 4 real representative examples from Garments2Look test set. QIE-2509 = Qwen-Image-Edit-2509, NB = Nano Banana, N Ref = Using a model image and multiple single garment images as input, 2 Ref = Using a model image and an OOTD image as input. A yellow box denotes the difference with the look image (GT). A black arrow indicates a distinct artifact boundary. Row 1: 4 items, 1 layer, no accessory. Row 2: 5 items, 2 layers, no accessory. Row 3: 8 items, 3 layers, 2 accessories. Row 4: 9 items, 3 layers, 3 accessories.
    }
    \vspace{-1.3em}
    \label{fig:compare}
\end{figure*}

\begin{figure}[t]
    \centering
    \includegraphics[width=\linewidth]{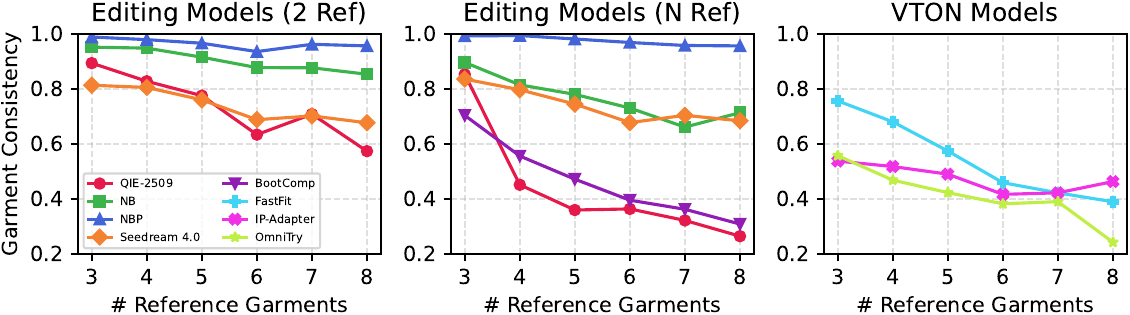}
    \vspace{-1.5em}
    \caption{Garment consistency with respect to the number of reference garment images, group by different types of model.}
    \label{fig:garment-acc}
    \vspace{-1.8em}
\end{figure}

As a newly proposed dataset and an advanced task in VTON family, Garments2Look carries clear practical value while introducing non-trivial challenges to existing approaches.
To clarify whether the task is overly simplistic for in-depth research or prohibitively difficult for feasible solutions, we evaluate state-of-the-art methods on our dataset quantitatively (\cref{table:garments2look-compare,fig:garment-acc}) and qualitatively (\cref{fig:compare}).
Our analysis focuses on the following four questions:

\textbf{Q1. How many items can be worn simultaneously?}
VTON models tied to fixed label sets struggle to handle a larger number of items. The iterative OmniTry~\cite{feng2025omnitry} paradigm tends to replace rather than add more layers of garment, often retaining only a single top. In contrast, general-purpose editing models are more flexible: using free-form multi-image inputs or an OOTD image as a unified reference facilitates stacking more items. Editing models such as Nano Banana~\cite{comanici2025gemini}, which leverage both multi-image inputs and OOTD image, successfully increase layering depth as complexity grows.
When the number of item exceeds 4, VTON models that were not or rarely trained on such cardinalities either drop items or render only an outer layer while omitting inner garments. Editing models, by comparison, exhibit better robustness to variable-length outfit lists. Within VTON paradigm, enabling variable-length inputs, or assembling inputs via an OOTD-style composition, appears to be a viable and promising direction for accommodating diverse, multi-item outfits.
These findings suggest that the input modality seen during training fundamentally caps the achievable outfit composition capacity.

% \usepackage{multirow}
\begin{table}[t]
\caption{Quantitative comparison on Garments2Look test set. We report results on classical VTON metrics and accuracy metrics judged by VLM. QIE-2509 = Qwen-Image-Edit-2509, NB = Nano Banana, NBP = Nano Banana Pro.}
\vspace{-0.8em}
\setlength{\tabcolsep}{2.2pt}
\resizebox{0.48\textwidth}{!}{
\begin{tabular}{lccccccc}
\toprule
Models & FID↓ & KID↓ & SSIM↑ & LPIPS↓ & Garment↑ & Layering↑ & Styling↑ \\ \midrule
\rowcolor{gray!20} \textit{VTON Models} & & & & & & & \\
IP-Adapter~\cite{ye2023ip-adapter} &
6.5511 & 6.7624 & 0.7960 & 0.1472 & 0.4950 & 0.6036 & 0.5693 \\
BootComp~\cite{Choi_2025_CVPR} & 
8.6301 & 8.9129 & 0.6110 & 0.3364 & 0.5369 & 0.3127 & 0.3547 \\
OmniTry~\cite{feng2025omnitry} & 
6.5559 & 10.0715 & 0.7244 & 0.2295 & 0.4612 & 0.1674 & 0.2609 \\
FastFit~\cite{chong2025fastfitacceleratingmultireferencevirtual} & 
3.5919 & 4.5827 & 0.8550 & 0.0956 & 0.6237 & 0.1305 & 0.3400 \\
\rowcolor{gray!20} \textit{Image Editing Models} & & & & & & & \\
GPT-4o~\cite{hurst2024gpt} (N Ref) & 
4.0478 & 2.8334 & 0.6648 & 0.2565 & 0.8355 & 0.7901 & 0.6401 \\
GPT-4o~\cite{hurst2024gpt} (2 Ref) & 
2.1496 & 1.4154 & 0.7577 & 0.1558 & 0.8921 & 0.8487 & 0.6943 \\
Seedream 4.0~\cite{seedream2025seedream} (N Ref) & 
4.1436 & 6.2291 & 0.7134 & 0.1831 & 0.7709 & 0.8312 & 0.6452 \\
Seedream 4.0~\cite{seedream2025seedream} (2 Ref) & 
3.7117 & 6.0362 & 0.7358 & 0.1683 & 0.7712 & 0.8808 & 0.6808 \\
QIE-2509~\cite{wu2025qwen} (N Ref) & 
33.1751 & 83.7688 & 0.5093 & 0.5360 & 0.5144 & 0.4716 & 0.5120 \\
QIE-2509~\cite{wu2025qwen} (2 Ref) & 
1.7562 & 1.0117 & 0.8111 & 0.0910 & 0.7940 & 0.8168 & 0.6729 \\
NB (N Ref)~\cite{comanici2025gemini} & 
1.1619 & \textbf{0.1628} & 0.8523 & 0.0663 & 0.8045 & 0.9250 & \textbf{0.7482} \\
NB (2 Ref)~\cite{comanici2025gemini} & 
\textbf{1.0412} & 0.2523 & \textbf{0.8583} & \textbf{0.0619} & 0.9253 & 0.8847 & 0.7386 \\
NBP (N Ref) & 
1.3182 & 0.4002 & 0.8167 & 0.0903 & \textbf{0.9836} & \textbf{0.9363} & 0.7356 \\
NBP (2 Ref) & 
1.3462 & 0.5041 & 0.8202 & 0.0863 & 0.9707 & 0.9317 & 0.7213 \\ \bottomrule
\end{tabular}
}
\vspace{-2.0em}
\label{table:garments2look-compare}
\end{table}

\textbf{Q2. How consistent are look images with references?}
\cref{fig:compare} indicates that degradation with more references is universal: (1) Shape Distortion: In almost all examples, the shapes of accessories (such as bags and earrings) showed significant deviations. (2) Texture Change: In Row 1 and 2, the text content/style on the clothing was altered (``PRADA'' and ``LOWEWE''), and in Row 3, the diagonal stripe texture of the sweater was not maintained consistently (GPT-4o and NB (2 Ref)). (3) Color Deviation: In Row 4, the color of the coat generated by BootComp and the knitwear generated by NB (N Ref) both deviated from the reference image. (4) Item Fusion: In Row 3, for BootComp, two upper garments that should have been independent are incorrectly merged. 
Moreover, with the same sample, the result by 2-reference strategy often outperforms N-reference strategy (See 2 Ref \vs N Ref results in~\cref{fig:compare,table:garments2look-compare,fig:garment-acc}). Treating the outfit as a holistic reference carries contextual co-occurrence and implicit relations, which current models leverage better than disparate references.

\textbf{Q3. How good is the overall try-on effect?}
This question targets the overall visual impression, evaluating global coordination and the handling of special dressing techniques. We observe inpainting artifacts, \eg, FastFit~\cite{chong2025fastfitacceleratingmultireferencevirtual} shows unnatural body-background transitions in \cref{fig:compare}, likely from a bias toward pure-white try-on backgrounds. Layered clothing remains problematic: although editing models can parse prompts specifying layers, control degrades as the number of reference items grows. Compared with single-garment cases (Row 1), layered scenarios (Rows 2–3) show weaker consistency, including missing text on the white inner tee (Row 2), incorrect blazer button counts (Row 3), distorted stripe density on the inner shirt (Row 4, QIE-2509 and NB (N Ref)), and inaccurate jacket length (Row 4, Seedream 4.0). Styling is also poorly controlled: In Row 4, the target look features an untucked mid-layer with only one-two buttons fastened; even with explicit prompts, most models produce tidy, tucked outfits, indicating a lack of fine-grained control for non-standard styling.

\textbf{Q4. Why the results on Garments2Look in~\cref{table:garments2look-compare} differ from the results on DressCode-MR~\cite{chong2025fastfitacceleratingmultireferencevirtual} reported in~\cref{table:dresscode-mr-compare}?}
We find that most of editing models outperform VTON models on our dataset. It can be attributed to two factors: 
(1) Garments2Look spans more diverse categories. Editing models are not restricted by item types and can accommodate more categories. In contrast, VTON models support only a limited set of categories, leaving many items (such as scarves, gloves, and brooches) untestable; these items are absent from DressCode-MR~\cite{chong2025fastfitacceleratingmultireferencevirtual} but included in Garments2Look. 
(2) Garments2Look contains more items and complex layering and styling. Editing models are more flexible for such settings, while SOTA VTON models focus on single-layer try-on, struggle with simultaneous multi-garment synthesis or only support simple instructions.

In summary, Garments2Look poses a substantial yet tractable challenge. It introduces a practical task and dataset that are difficult enough to stress current methods while remaining fertile for meaningful research progress.

\subsection{What new insights does it bring?}

Existing SOTA VTON methods perform poorly on Garments2Look, with few completing the full try-on pipeline. We identify three key causes: (1) paradigm shifts caused by large-scale fashion inventory; (2) hierarchical dependencies among layered garments; (3) high sensitivity to fine-grained design details. These reveal fundamental limitations in purely visual frameworks.

To address these constraints, Garments2Look introduces rich structured textual annotations, a dimension largely absent in traditional VTON datasets. 
Unlike previous methods such as BootComp~\cite{Choi_2025_CVPR} and OmniTry~\cite{feng2025omnitry}, which treat text merely as independent signals, our dataset bridges the gap between visual inputs and complex outfit semantics. 
As detailed in~\cref{sec:Synthesis}, we provide high-fidelity annotations aligned with image inputs, encompassing item-level and outfit-level descriptions, including garment categories, garment descriptions, outfit overall descriptions, layering logic, styling techniques, model attributes, \etc. This structured approach aligns with the evolving landscape of multi-modal learning~\cite{kim2025promptdresser, guo2025enhancing, yu2026modality}, providing the necessary semantic guidance for complex outfit synthesis.

\begin{table}[t]
\caption{Study of the detail level of the text prompt on sampled data of Garments2Look test dataset, generated by Nano Banana.}
\vspace{-0.6em}
\resizebox{0.48\textwidth}{!}{
\begin{tabular}{lcccc}
\toprule
Textual Input Type & FID↓ & KID↓ & SSIM↑ & LPIPS↓ \\
\midrule
Item Category & 23.272 & 1.271 & 0.817 & 0.141 \\
Item Category + Outfit Overall & 22.135 & 0.752 & 0.810 & 0.155 \\
Item Category + Layering \& Styling &  21.831 & \underline{0.733} & 0.814 & 0.148 \\
Item Category + Shape \& Pose & \underline{21.783}  & 0.750 & \underline{0.823} & \underline{0.133} \\
Item Category + Shape \& Pose + Layering \& Styling &  \textbf{21.545} & \textbf{0.641} & \textbf{0.825} & \textbf{0.131} \\
\bottomrule
\end{tabular}
}
\vspace{-0.6em}
\label{table:prompt-ablation}
\end{table}

\begin{figure}[t]
    \centering
    \includegraphics[width=1\linewidth]{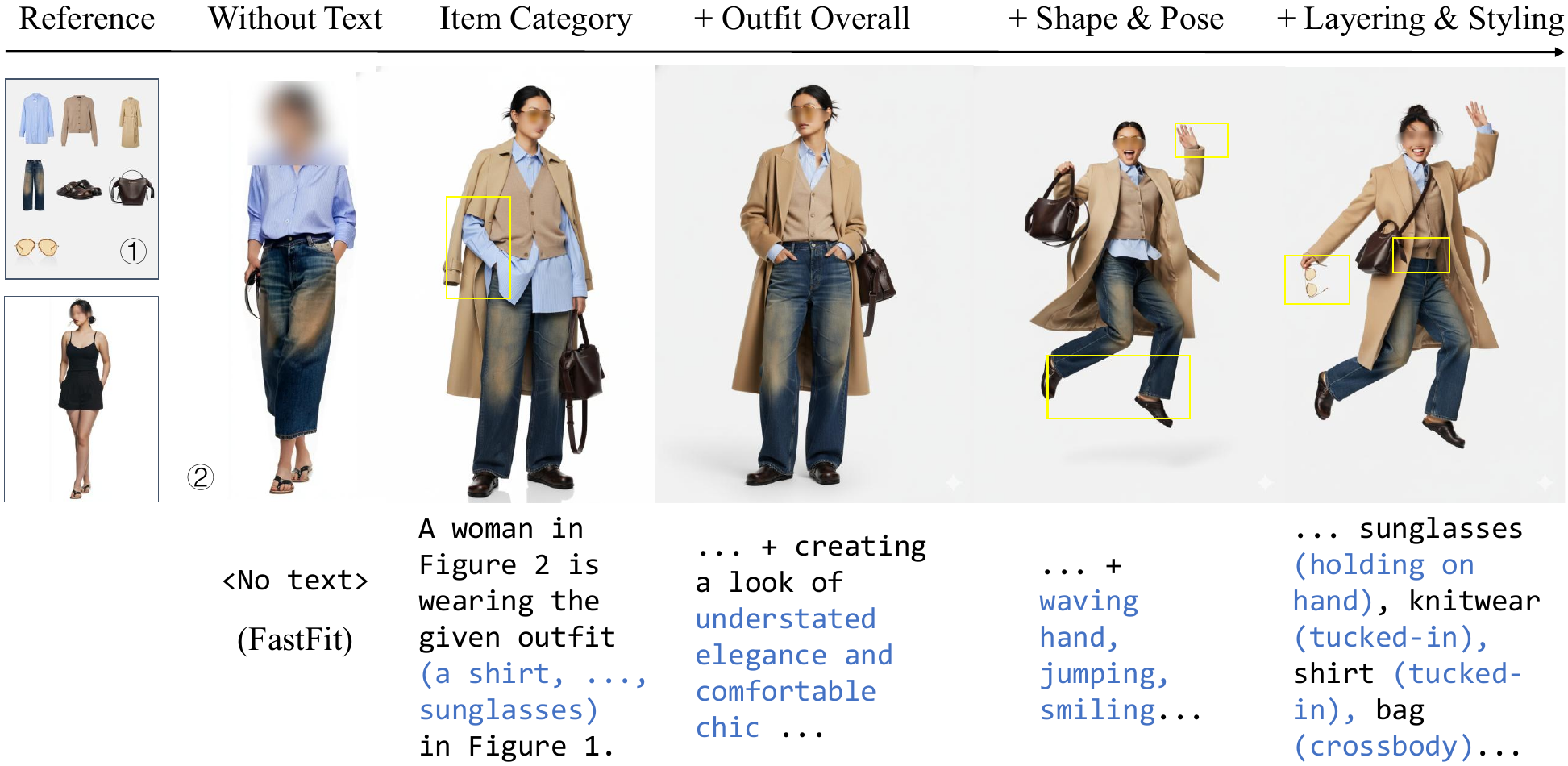}
    \vspace{-2em}
    \caption{Visualization of the contribution of text modality.}
    \vspace{-1.8em}
    \label{fig:text-modality}
\end{figure}

Our experiments confirm the effectiveness of text guidance in complex VTON scenarios.
In~\cref{table:prompt-ablation}, We conduct an ablation study to evaluate the influence of textual granularity.
While item types are provided as basic guidance (Row 1), integrating outfit-level information (Row 2) yields measurable gains in FID and KID. As we incrementally compose finer-grained attributes (Rows 3-5), the model achieves a significant performance leap across all metrics.
This trend shows that the textual modalities in Garments2Look offer essential guidance that synergizes with visual features to solve outfit-level VTON task.
A representative example in~\cref{fig:text-modality} further substantiates these observations. 
By employing our comprehensive and unique text annotations, the model generates more stylish and higher-quality fashion images. The last two examples also illustrate improved controllability over pose and style, highlighting the flexible generation potential enabled by Garments2Look.

These results suggest that beyond-vision features are essential for scaling VTON to realistic, multi-layered wardrobes and attribute-sensitive fashion applications.
% %

\section{Conclusions}

This paper addresses the critical gap of outfit-level VTON in existing datasets, which lack support for multi-garment layering, accessory integration, and fine-grained layering and styling annotations. 
We introduce Garments2Look, the first large-scale multimodal dataset for outfit-level VTON, comprising 80K high-fidelity pairs with structural annotations. 
Our evaluation of SOTA methods reveals substantial shortcomings in generating outfit-level results, underscoring the new challenges posed by our setting and highlighting insights that point to the potential of this direction.
Future work will focus on designing more suitable task-specific metrics, and exploring models that fuse visual and textual cues for end-to-end outfit-level VTON with precise details.

\section*{Acknowledge}
The work described in this paper was substantially supported by a grant from the Research Grants Council of the Hong Kong Special Administrative Region, China (Project No. PolyU/RGC Project  25211424) and partially supported by a grant from PolyU University Start-Up Fund (Project No. P0047675).

{
    \small
    \bibliographystyle{ieeenat_fullname}
    \bibliography{main}
}

\clearpage
\newpage
\appendix 
\setcounter{figure}{0}
\setcounter{table}{0}
\renewcommand\thesection{\Alph{section}}
\renewcommand\thetable{S\arabic{table}}
\renewcommand\thefigure{S\arabic{figure}}

\twocolumn[{%
\renewcommand\twocolumn[1][]{#1}%
\maketitlesupplementary
\includegraphics[width=1.0\linewidth]{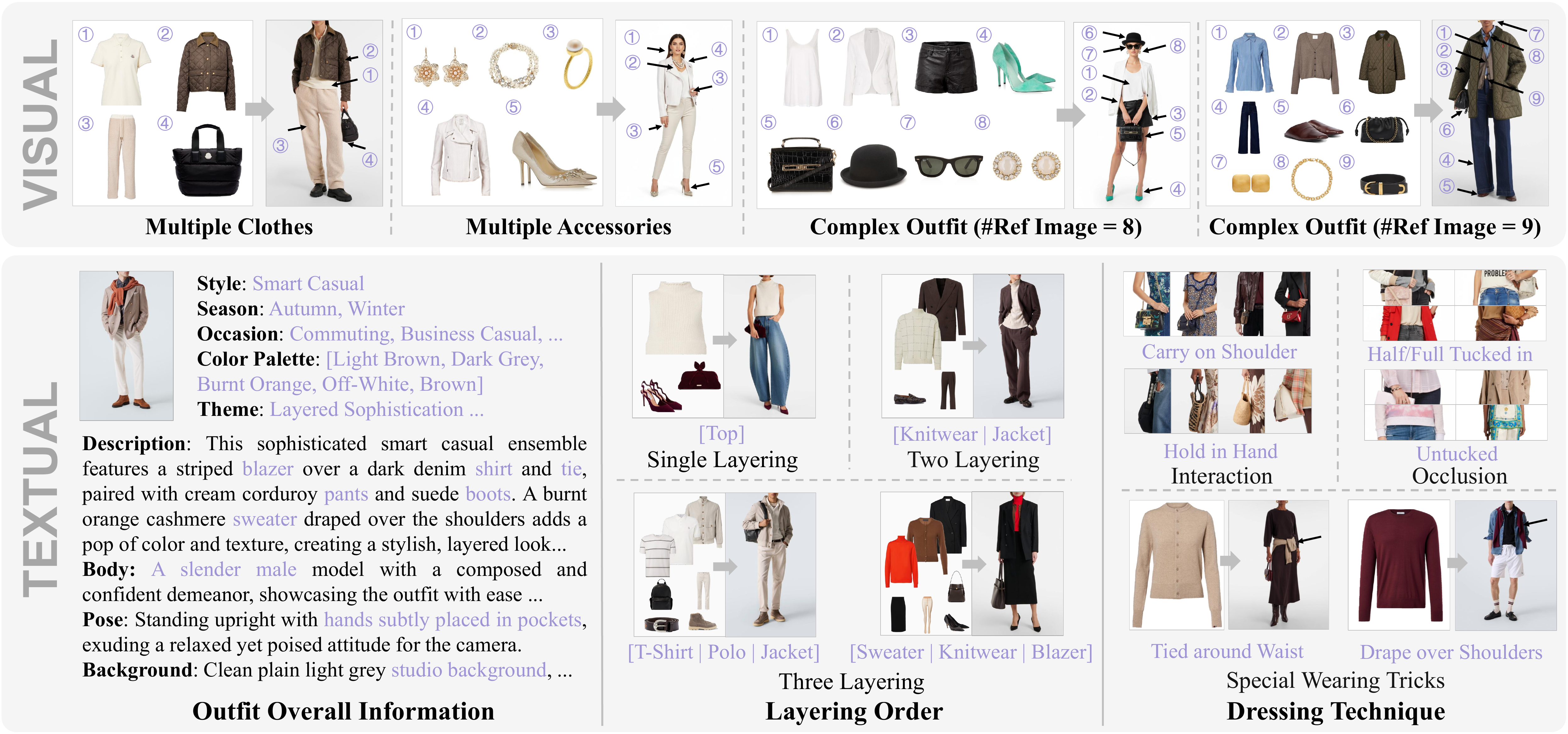}
\captionof{figure}{Example data in our outfit-level VTON dataset, Garments2Look.}
\vspace{1em}
\label{fig:teaser}
}]

\section{Dataset Details}
\label{sec:supp:details}

\subsection{Visual Samples in Garments2Look}

We showcase additional example data from our proposed dataset in~\cref{fig:teaser}. This dataset provides high-quality, high-precision, and highly diverse samples of complex and complete clothing combinations.

In terms of visual dimensions, we take a real-world application perspective, covering more complex virtual try-on scenarios than previous datasets. This includes everything from basic items to rich accessories, from single-layer outfits to multi-layered looks, and from single-style dressing to multiple styling options. For example, a simple combination of a T-shirt, shorts, shoes, and a bag often involves adding accessories like earrings, bracelets, hats, or glasses as focal points in real-world scenarios. Alternatively, a thin knitted cardigan can be layered over it to create a sense of depth. There are also opportunities to tie a scarf in a knot at the chest or around the waist. Beyond the increased dimensionality to fulfil practical usage, we also focused on ensuring data quality. From the item images or outfits collection, outfit composition generation, to try-on synthesis, everything was reviewed by fashion experts.

Furthermore, recognizing the importance of the text modality—and that certain information (\eg, styling tips) is better conveyed through language—unlike previous try-on datasets, we provide rich and detailed textual descriptions for all samples. These include item categories, fine-grained attributes (style, season, occasion, color, theme, \etc), overall descriptions of multi-item outfits, the order and manner in which items are worn, and factors that directly affect fit, such as the model’s body shape and posture.

We expect this dataset will strongly support deeper research and broader applications in virtual try-on.

\subsection{Data Division}

As shown in~\cref{tab:data-ratio}, we constructed a test set of $\sim$1K samples for  evaluation. This set was stratified sampled from the full 80K samples to ensure high representativeness, and authenticity of data sources. All remaining samples were designated as the training set.

\begin{table}[tbh]
\centering
\small
\captionof{table}{Division of Garments2Look.}
\vspace{-0.6em}
\setlength{\tabcolsep}{4pt}
% \resizebox{1\linewidth}{!}{%
\begin{tabular}{lccc}
\toprule
Type & Train & Test & Total \\
\midrule
Real & 39819 & 402 & 40222 \\
Synthetic & 39278 & 541 & 39819 \\
Total & 79097 & 944 & 80041 \\
\bottomrule
\end{tabular}%
% }
\label{tab:data-ratio}
\end{table}

\section{Data Process}

\subsection{Outfit Synthesis}
\label{supp:sec:outfit-synthesis}

\textbf{Style Guidance}: First, the style categories are determined based on popular classifications obtained through web retrieval. Next, the initial drafts are generated by Gemini-2.5-Flash. Finally, the content is revised by three senior fashion experts to ensure its professionalism and accuracy.
An example of a style guide (Minimalist Style) is provided in~\cref{tab:minimalist_style_guide_prompt}.

\noindent\textbf{Primary Fashion Styles}: American Vintage, Androgynous, Beach, Bohemian, Business Casual, Casual, Classic, Comfort, Cottagecore, Country, Cowboy, Eclectic, Elegant, Formal, French chic, Girly, Glamorous, Gorpcore, Gothic, Hip-Hop, Hong Kong Vintage, Kawaii, Lagenlook, Military, Minimalist, Mori Girl, Moto, Old Money, Playful, Preppy, Punk, Quirky, Romantic, Spicy, Sporty, Street, Workwear, Y2K.

\noindent\textbf{Retrieval Sampling}: We introduce an inverse frequency-based reweighted retrieval sampling mechanism. We select the $N$ candidate items $\{x_1, \dots, x_N\}$ with the highest similarity from the retrieval results. Subsequently, we count the historical occurrence of each item $x_i$ in current outfit dataset, denoted as $c_i$, and use this to define its inverse frequency weight $w_i = (c_{\max} - c_i) + 1$, where $c_{\max} = \max_{j=1}^{N} (c_j)$. During the sampling process, the probability $f(c_i)$ of item $x_i$ being selected is determined by the ratio of its weight to the total weight:
$f(c_i) = \frac{w_i}{\sum_{j=1}^{N} w_j}$.
This design encourages the model to select items with low historical occurrence frequency by assigning lower weights to historically high-frequency items.

\subsection{Look Synthesis}
\label{sec:supp:look-synthesis}

\noindent\textbf{Text Annotation of Look Image}: 
We generate rich textual annotations describing each try-on images. These annotations cover: (i) an overall description of the try-on image; (ii) how each item is worn, including layering order and styling techniques; and (iii) details about the model, such as body shape and pose, and backgrounds. Gemini-2.5-Flash generated all text annotations, and they were subsequently reviewed by fashion experts.

\begin{figure}[t]
    \centering
    \includegraphics[width=\linewidth]{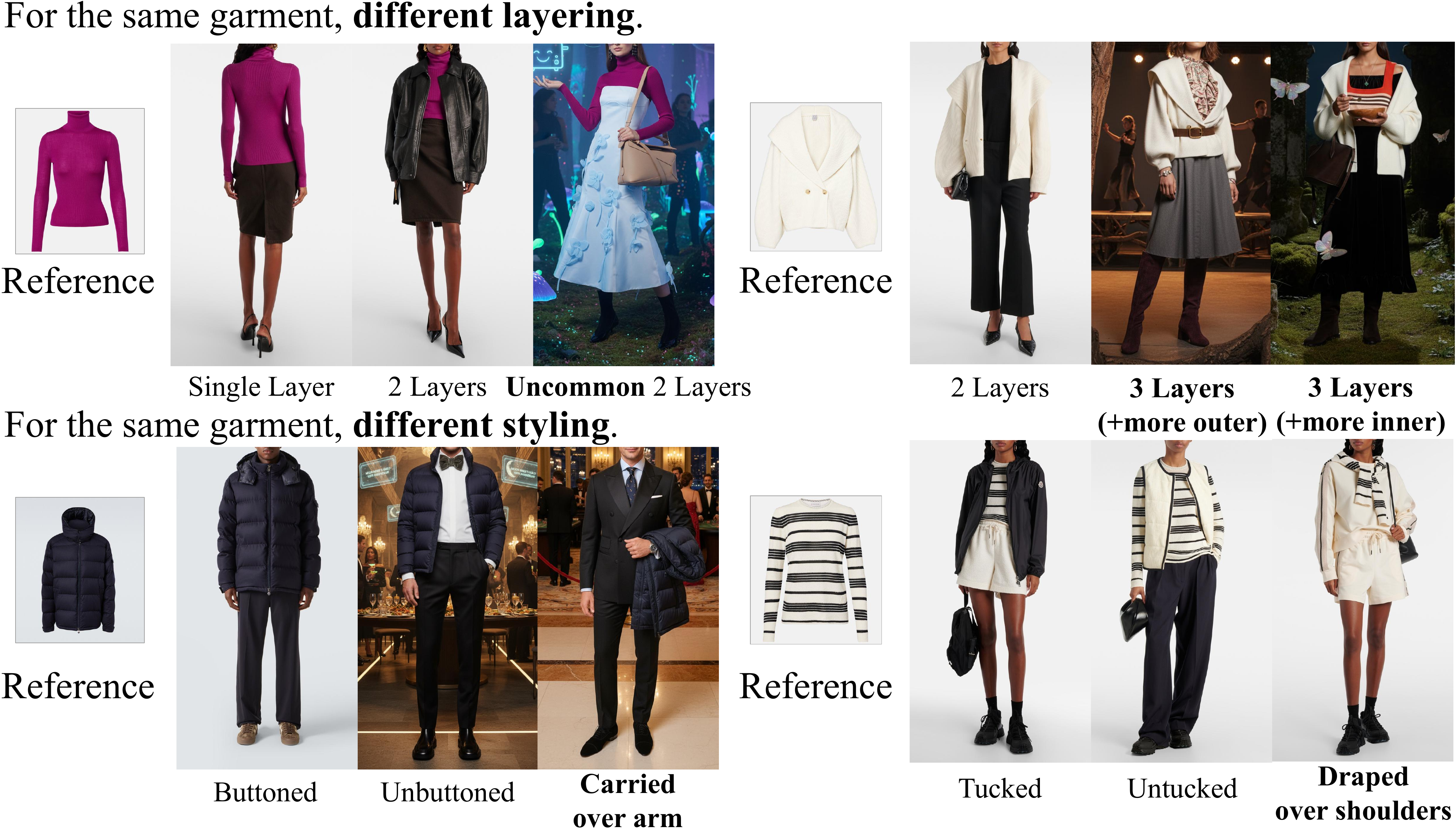}
    \caption{More layering \& styling in Garments2Look.}
    \label{fig:supp-layering-styling}
\end{figure}

\noindent\textbf{Layering \& Styling.}
As shown in~\cref{fig:supp-layering-styling}, our dataset captures the versatility of individual garments by including multiple outfit instances for the same piece. This design enables the same garment to appear in different layering configurations (\eg, as a single base layer, paired with an outer jacket, or combined with an unconventional dress overlay) and to be styled in distinct ways (\eg, carried over arm, or draped over the shoulders). Such diversity is valuable for advancing research on in-the-wild virtual try-on and garment layering order editing, as it provides rich examples of how garments interact with other pieces, mirroring the complexity of real-world fashion presentation.

\subsection{Data Filtering}
\label{supp:sec:filtering}

\textbf{Primary Categories List}: Activewear, Bag Accessories, Bags, Beachwear, Belts, Bracelets, Brooches, Coats, Cuff Links, Tie Clips, Dresses, Earrings, Glasses, Gloves, Hair Accessories, Hats, Jackets, Jeans, Jumpsuits, Knitwear, Lingerie, Necklaces, Pants, Rings, Scarves, Shirts, Shoes, Shorts, Ski Goggles, Skirts, Skiwear, Socks, Tights, Suits, Sunglasses, Ties, Bow Ties, Tops, T-shirts, and Watches.

\noindent\textbf{Categories Correction}: 
We manually corrected the category structure in the original metadata.
We merge semantically similar subcategories (\eg, consolidating ``Sportswear Jacket'' and ``Suit Jacket'' under the ``Jacket'' category), and exclude irrelevant categories (\eg, home goods, phone cases, \etc).
This process ensures subsequent outfit and look synthesis are focused on wearable garments. 

\noindent\textbf{Fashion Expert Review Process}: 
To guarantee the quality of the synthetic data, we implemented a rigorous quality control pipeline involving 13 human fashion experts for review: (1) Guideline Development, establishing a rubric for quality and styling; (2) Pilot Study, using a 2\% batch for double-blind labeling to refine criteria; and (3) Mass Annotation, employing cross-validation with a senior expert arbitrating any discrepancies to ensure consensus.

\section{More Experiments}
\label{sec:more-exp}

\subsection{Split Results}
\label{sec:split-result}

In~\cref{tab:split-data}, we presents the spilit results; Nano Banana series maintain SOTA performance across both test subsets, outperforming all baselines.

\subsection{Influence of Explicit Pose Control}
\label{sec:pose-condition}

To investigate this, we sampled test data and utilized 3 references for image generation on our test set: the masked model image, the skeleton image, and the OOTD image.
Based on the experimental results shown in~\cref{table:pose-condition-3-refer}, we observed that adding the skeleton image as an additional reference did not significantly improve the model's metrics; instead, performance slightly decreased. This suggests that general image editing models may struggle to effectively integrate and disambiguate ``different type'' of reference information (pose structure \vs object reference) simultaneously. When we introduce the skeleton image as a third input and treat it as equally important as the garment item image, the model may fail to correctly interpret the skeleton's role as a pose constraint. Rather, this extra input can introduce information redundancy or confusion, interfere with the model's editing capability, and ultimately lead to a drop in metrics. This contrasts with models specifically designed for VTON, which can effectively leverage skeleton information.

\subsection{Results after Fine-Tuning}
\begin{figure}[t]
    \centering
    \includegraphics[width=\linewidth]{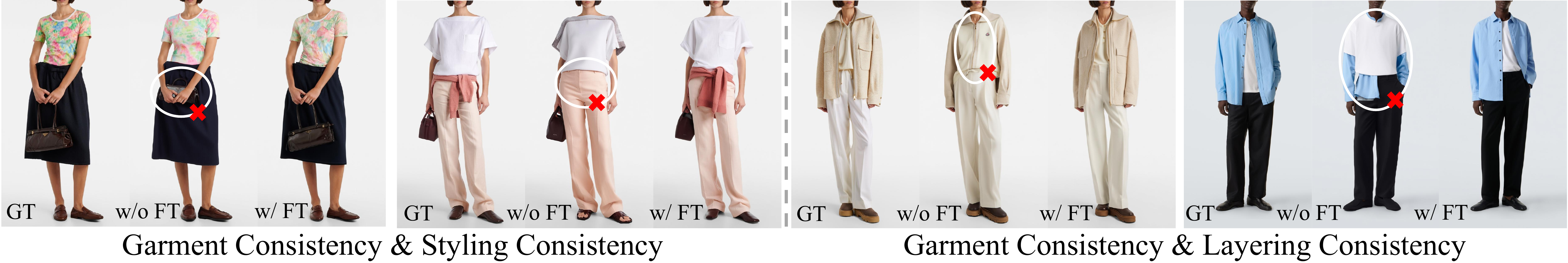}
    \vspace{-1.4em}
    \caption{Qualitative results after fine-tuning QIE-2509 on 10K random sampled data from Garments2Look training set.}
    \label{fig:qwen-image-edit-2509-ft}
\end{figure}

All results in the main paper are not fine-tuned. We add results fine-tuning Qwen-Image-Edit-2509 on our training set (randomly sampled 10K due to the time constraint). As shown in~\cref{tab:split-data}, \cref{fig:qwen-image-edit-2509-ft} and~\cref{fig:layer-user-study} (QIE-2509+FT), the quantitative and qualitative results, and user study all demonstrate improvements.

\begin{table}[t]
\centering
\small
\caption{\footnotesize Separated evaluation results on Garments2Look test set (golden/synthetic).}
\vspace{-0.9em}
\setlength{\tabcolsep}{2.2pt}
\resizebox{\linewidth}{!}{%
\begin{tabular}{l *{6}{c@{\,/\,}c}}
\toprule
Method & 
\multicolumn{2}{c}{KID$\downarrow$} & 
\multicolumn{2}{c}{SSIM$\uparrow$} & 
\multicolumn{2}{c}{LPIPS$\downarrow$} &
\multicolumn{2}{c}{Garment Acc.$\uparrow$} & 
\multicolumn{2}{c}{Layering Acc.$\uparrow$} & 
\multicolumn{2}{c}{Styling Acc.$\uparrow$} \\
\midrule
IP-Adapter                  & 8.3390  & 8.0255  & 0.8588 & 0.7492 & 0.1076 & 0.1768 & 0.5224 & 0.4747 & 0.6424 & 0.5948 & 0.6001 & 0.5506 \\
BootComp                    & 13.5787 & 14.3927 & 0.7702 & 0.4925 & 0.2399 & 0.4083 & 0.6354 & 0.4636 & 0.3059 & 0.3142 & 0.4374 & 0.3048 \\
OmniTry                     & 9.5100  & 14.8091 & 0.8036 & 0.6654 & 0.2011 & 0.2507 & 0.5292 & 0.4105 & 0.1371 & 0.1743 & 0.3576 & 0.2024 \\
FastFit                     & 4.6095  & 6.1260  & 0.8559 & 0.8544 & 0.0968 & 0.0948 & 0.7189 & 0.5527 & 0.1266 & 0.1313 & 0.4727 & 0.2599 \\
\midrule
GPT-4o (2 Ref)              & 3.7019  & 0.8719  & 0.8179 & 0.7128 & 0.1378 & 0.1693 & 0.9320 & 0.8624 & 0.8987 & 0.8374 & 0.7044 & 0.6881 \\
Seedream 4.0 (2 Ref)        & 12.4870 & 13.7404 & 0.8126 & 0.6786 & 0.1740 & 0.1640 & 0.7805 & 0.7643 & 0.8513 & 0.8875 & 0.6286 & 0.7124 \\
NB (N Ref)                  & 0.6837  & 0.9384  & 0.8745 & 0.8357 & 0.0753 & 0.0596 & 0.8562 & 0.7659 & 0.9272 & 0.9245 & 0.7125 & 0.7697 \\
NB (2 Ref)                  & 0.4981  & 1.1012  & 0.8769 & 0.8445 & 0.0699 & 0.0559 & 0.9503 & 0.9066 & 0.8703 & 0.8880 & 0.6844 & 0.7714 \\
NB Pro (N Ref)              & 0.6596  & 0.2797  & 0.8072 & 0.8238 & 0.1156 & 0.0714 & 0.9808 & 0.9783 & 0.9483 & 0.9336 & 0.6966 & 0.7592 \\
NB Pro (2 Ref)              & 1.0233  & 0.1314  & 0.8112 & 0.8269 & 0.1108 & 0.0680 & 0.9796 & 0.9641 & 0.9082 & 0.9370 & 0.6712 & 0.7516 \\
\midrule
QIE-2509 (2 Ref)        & 0.4754  & 1.6250  & 0.8874 & 0.7543 & 0.0609 & 0.1135 & 0.8704 & 0.7371 & 0.8639 & 0.8061 & 0.7074 & 0.6520 \\
QIE-2509 (2 Ref) + FT    & 0.1116  & 0.2199  & 0.8997 & 0.8674 & 0.0481 & 0.0563 & 0.9410 & 0.8944 & 0.9873 & 0.9539 & 0.8453 & 0.8679 \\
\bottomrule
\end{tabular}%
}
\label{tab:split-data}
\end{table}

% \usepackage{multirow}
\begin{table}[t]
\caption{Ablation experiment for using the skeleton image (pose condition) as an addition reference on image editing models. The format is ``2 Ref / 3 Ref  (3 Ref $-$ 2 Ref Delta Value)''.}
\vspace{-0.6em}
\setlength{\tabcolsep}{1pt}
\resizebox{0.49\textwidth}{!}{
\begin{tabular}{lccccc}
\toprule
Models & FID↓ & KID↓ & SSIM↑ & LPIPS↓ \\
\midrule
Seedream 4.0~\cite{seedream2025seedream} 
& 34.294 / 39.810 (+5.516) & 10.446 / 13.310 (+2.864) & 0.757 / 0.729 (-0.028) & 0.335 / 0.349 (+0.014) \\
QIE-2509~\cite{wu2025qwen} 
& 29.030 / 32.773 (+3.743) & 7.142 / 9.654 (+2.512) & \textbf{0.827} / 0.825 (-0.002) & 0.116 / \textbf{0.114} (-0.002) \\
NB~\cite{comanici2025gemini} 
& \textbf{21.545} / 21.484 (-0.061) & \textbf{0.641} / 0.867 (+0.226) & 0.825 / 0.824 (-0.001) & 0.131 / 0.131 (+0.000) \\
NBP
& 24.293 / 24.469 (+0.176) & 2.160 / 2.217 (+0.057) & 0.797 / 0.800 (+0.003) & 0.174 / 0.179 (+0.005) \\\bottomrule
\end{tabular}
}
\label{table:pose-condition-3-refer}
\vspace{-0.2em}
\end{table}

\subsection{User study on Layering}
\begin{figure}
    \centering
    \includegraphics[width=0.6\linewidth]{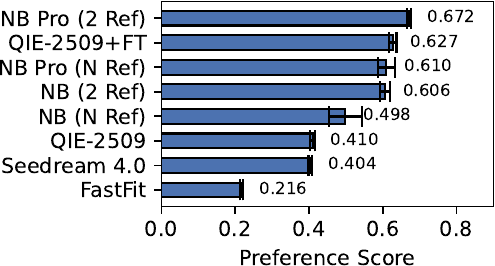}
    \caption{User study on layering accuracy.}
    \label{fig:layer-user-study}
\end{figure}

We conducted a user study focused on layering quality, specifically evaluating aspects such as occlusion rationality and layering accuracy. As shown in \cref{fig:layer-user-study}, the results reveal significant performance disparities among different methods, reflecting their varying capacities to leverage annotations.  FastFit, limited by its single-layer processing capability, received the lowest preference scores. Notably, after fine-tuning on a small subset of our data, QIE-2509’s performance improved significantly from 0.41 to 0.627. This highlights the intrinsic value of our dataset and demonstrates that models can effectively digest and utilize the rich information it provides.

\subsection{More Examples and Results}
\begin{figure*}[t]
    \centering
    \includegraphics[width=0.94\linewidth]{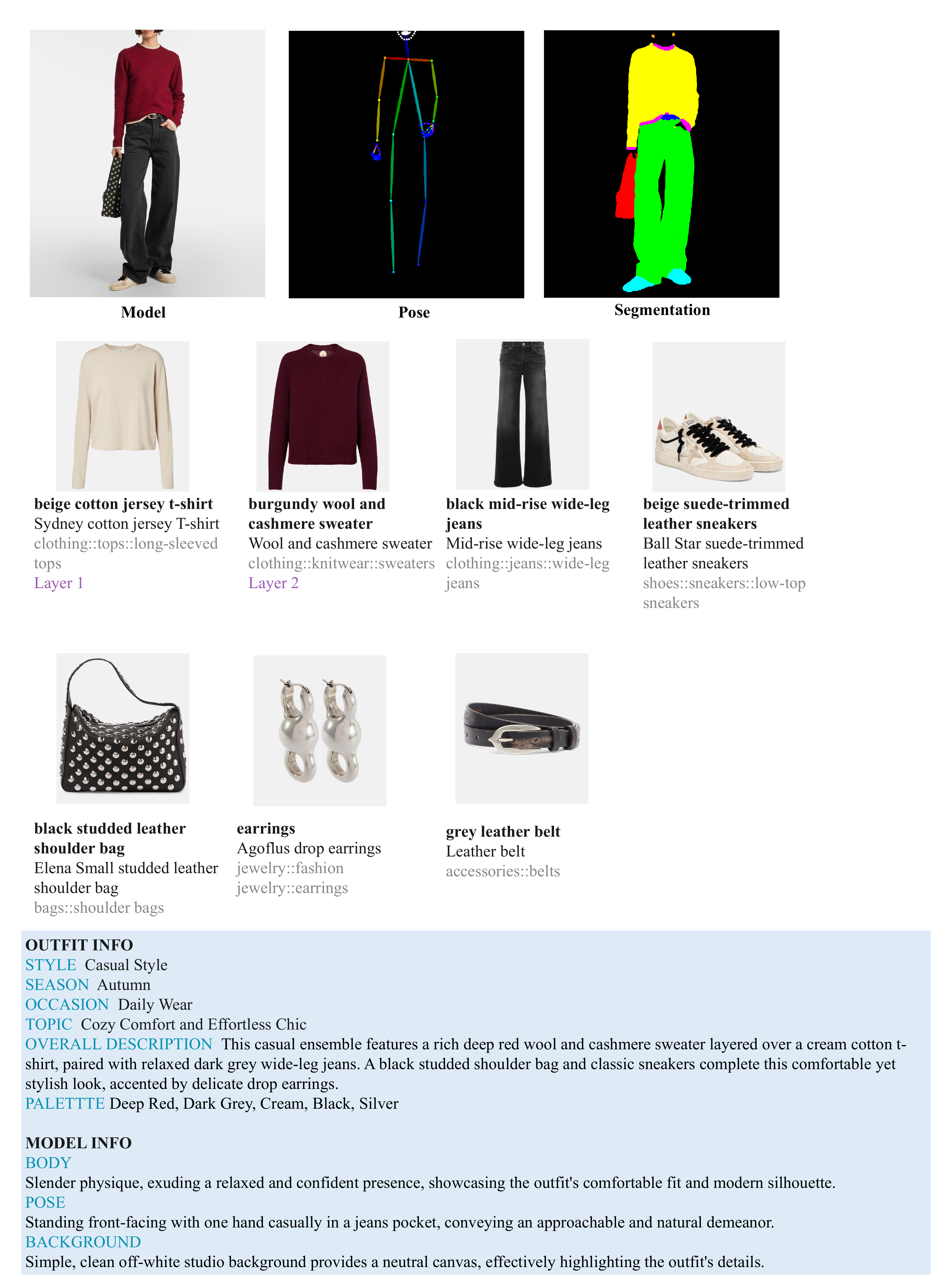}
    \vspace{-1.2em}
    \caption{More example data in Garments2Look (1/5).}
    \label{fig:supp-more-example-data-1}
\end{figure*}

\begin{figure*}[t]
    \centering
    \includegraphics[width=0.94\linewidth]{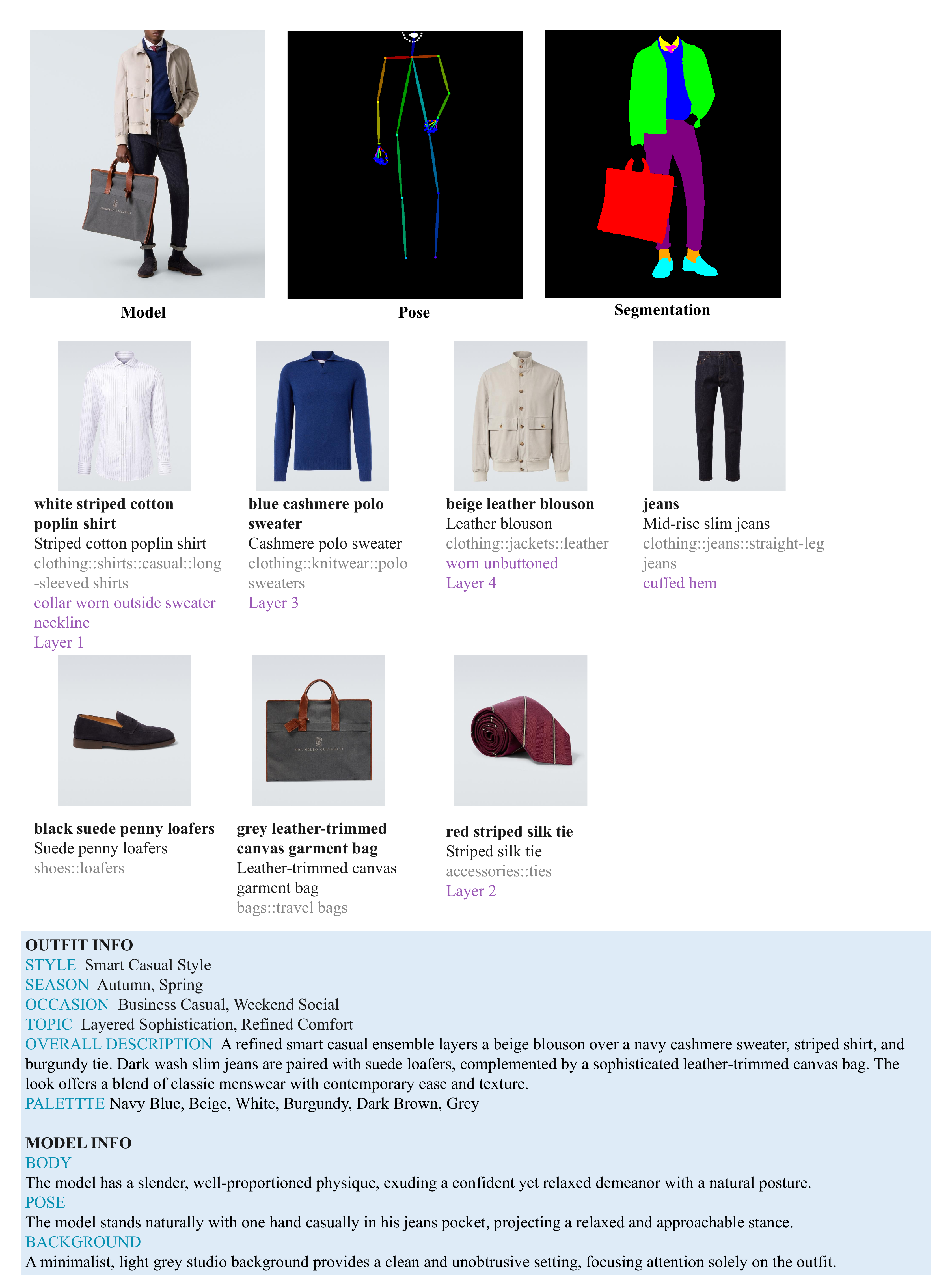}
    \vspace{-1.2em}
    \caption{More example data in Garments2Look (2/5).}
    \label{fig:supp-more-example-data-2}
\end{figure*}

\begin{figure*}[t]
    \centering
    \includegraphics[width=0.94\linewidth]{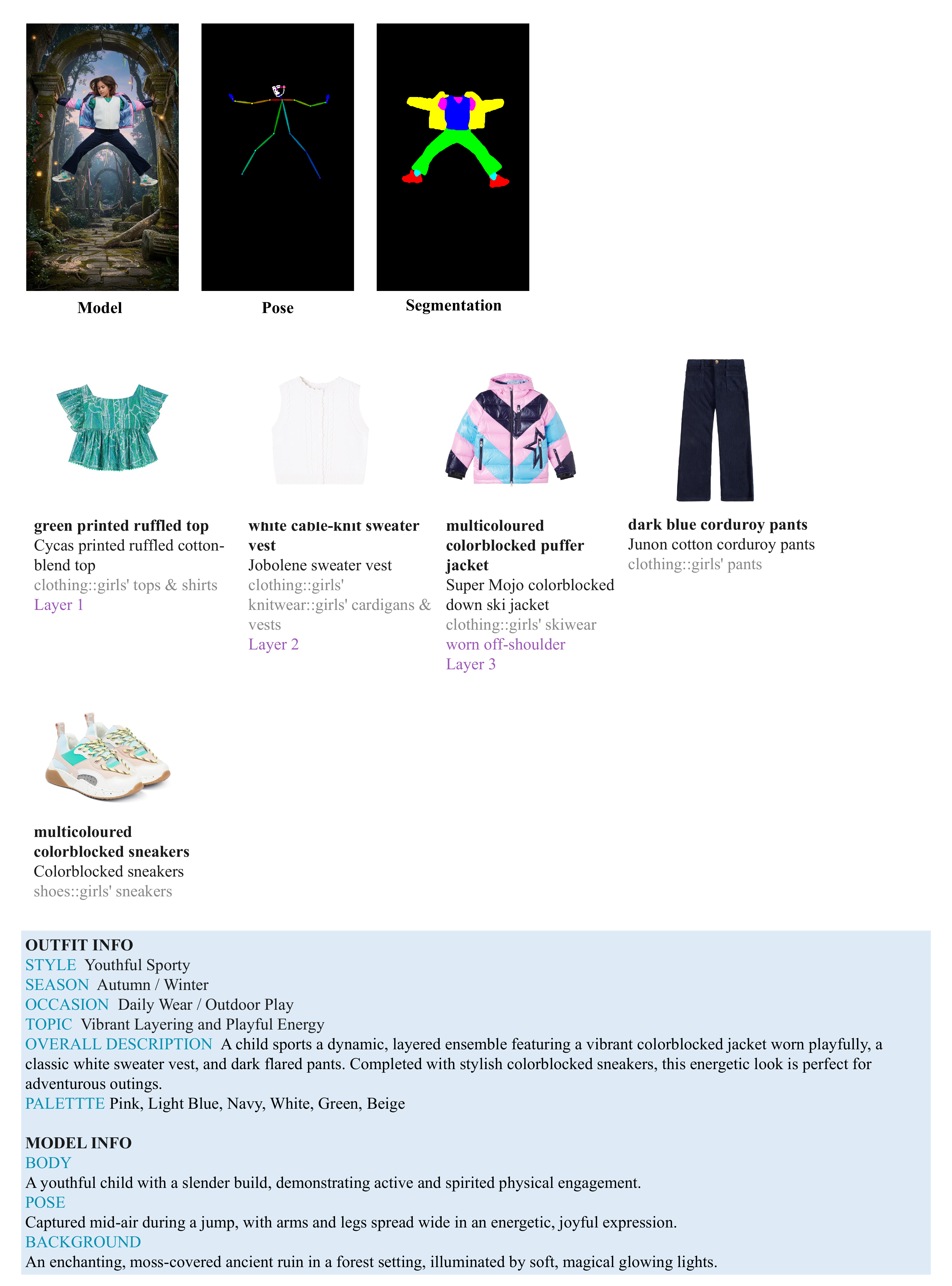}
    \vspace{-1.2em}
    \caption{More example data in Garments2Look (3/5).}
    \label{fig:supp-more-example-data-3}
\end{figure*}

\begin{figure*}[t]
    \centering
    \includegraphics[width=0.94\linewidth]{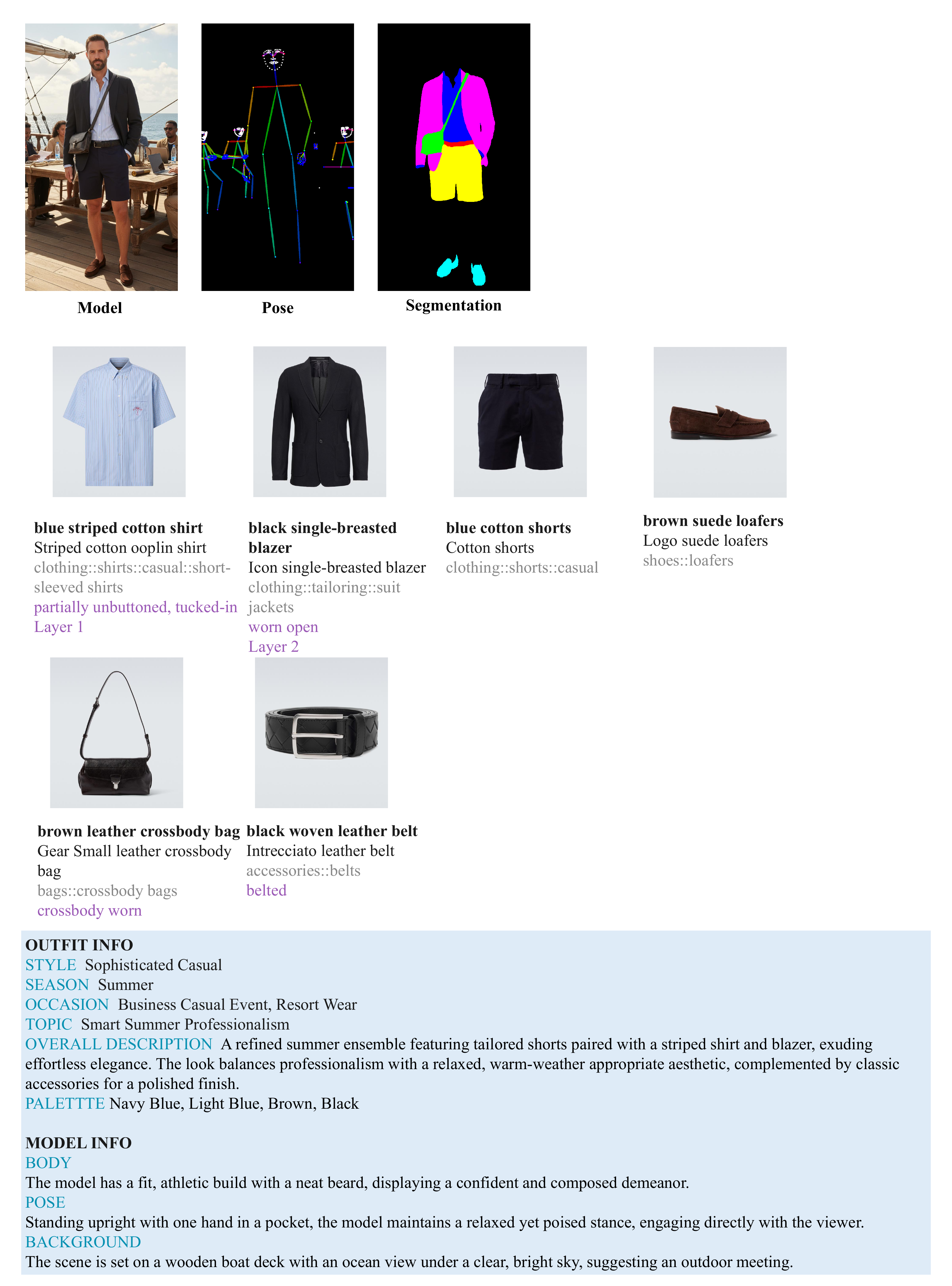}
    \vspace{-1.2em}
    \caption{More example data in Garments2Look (4/5).}
    \label{fig:supp-more-example-data-4}
\end{figure*}

\begin{figure*}[t]
    \centering
    \includegraphics[width=0.94\linewidth]{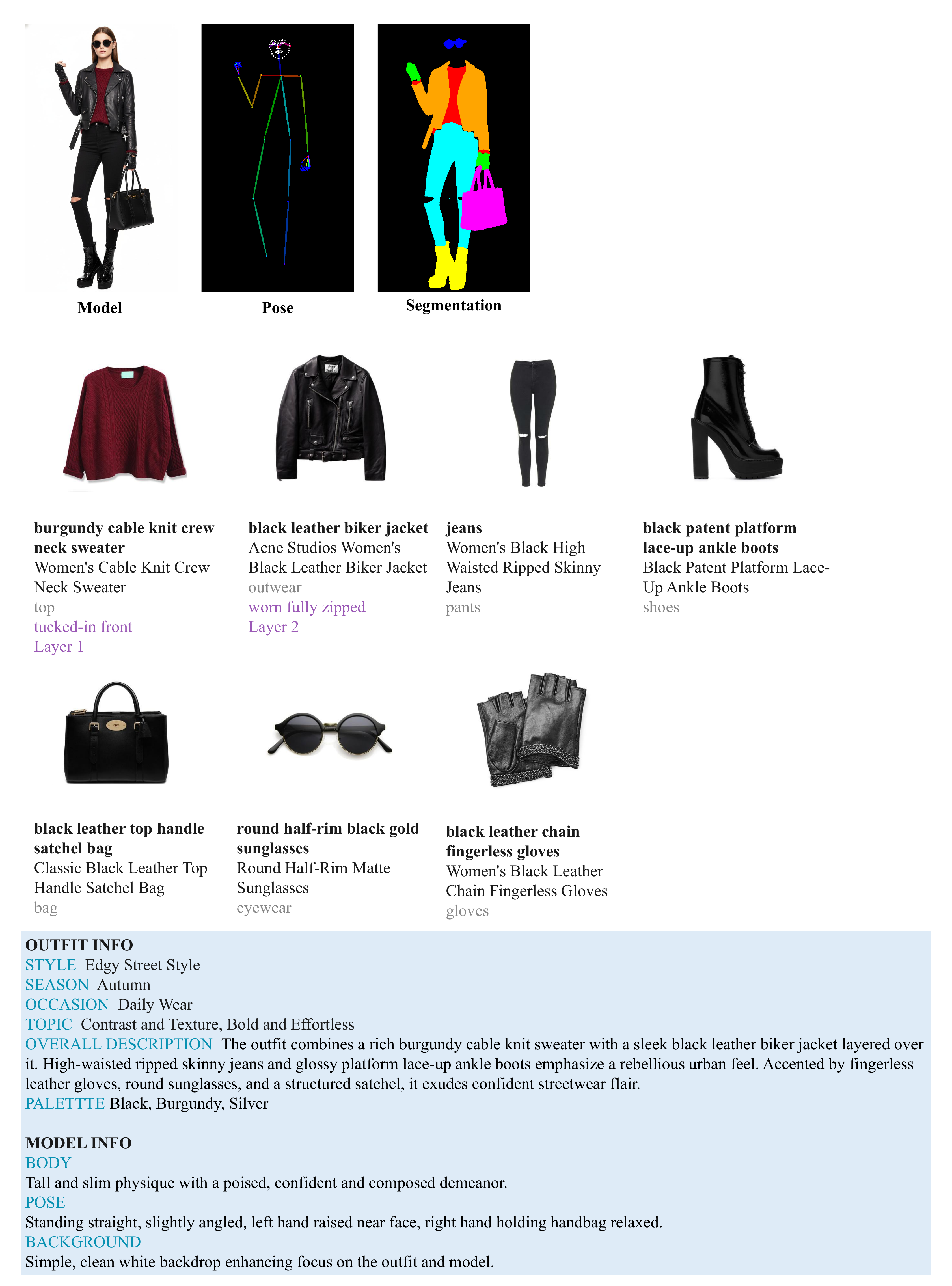}
    \vspace{-1.2em}
    \caption{More example data in Garments2Look (5/5).}
    \label{fig:supp-more-example-data-5}
\end{figure*}

\begin{figure*}[t]
    \centering
    \includegraphics[width=0.94\linewidth]{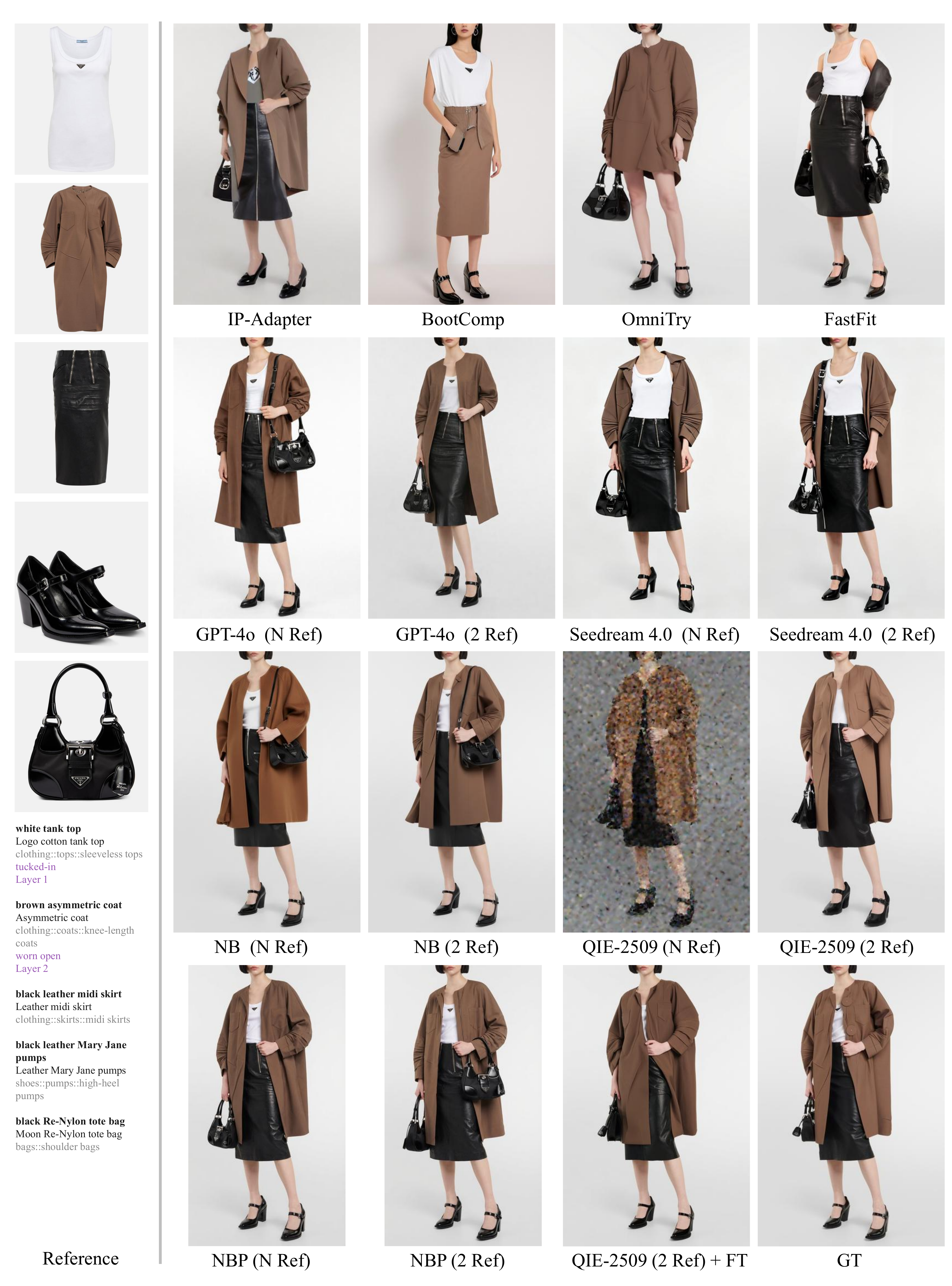}
    \vspace{-1.2em}
    \caption{More comparison results on Garments2Look test set (1/5).}
    \label{fig:supp-more-comparison-results-1}
\end{figure*}

\begin{figure*}[t]
    \centering
    \includegraphics[width=0.94\linewidth]{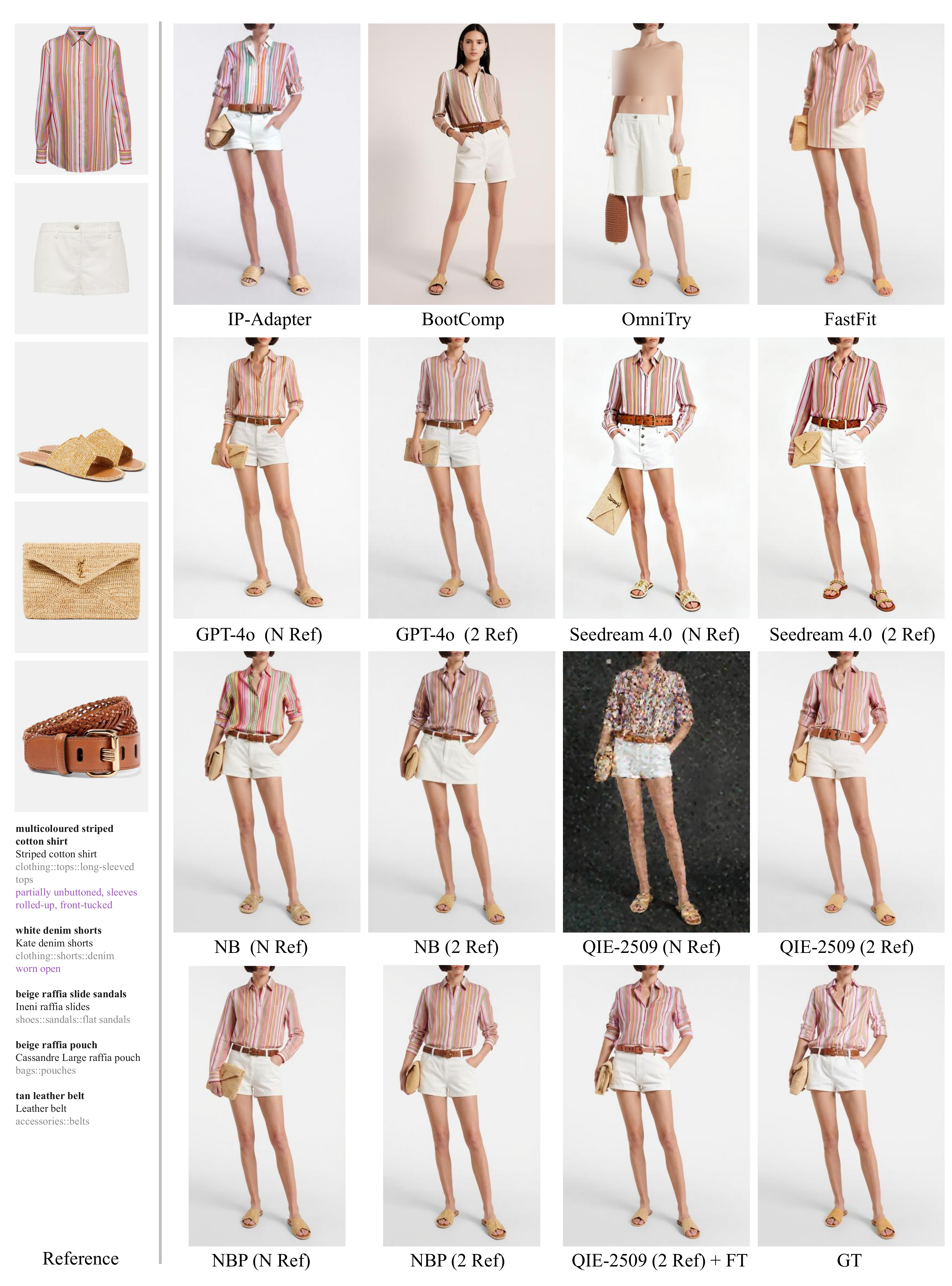}
    \vspace{-1.2em}
    \caption{More comparison results on Garments2Look test set (2/5).}
    \label{fig:supp-more-comparison-results-2}
\end{figure*}

\begin{figure*}[t]
    \centering
    \includegraphics[width=0.94\linewidth]{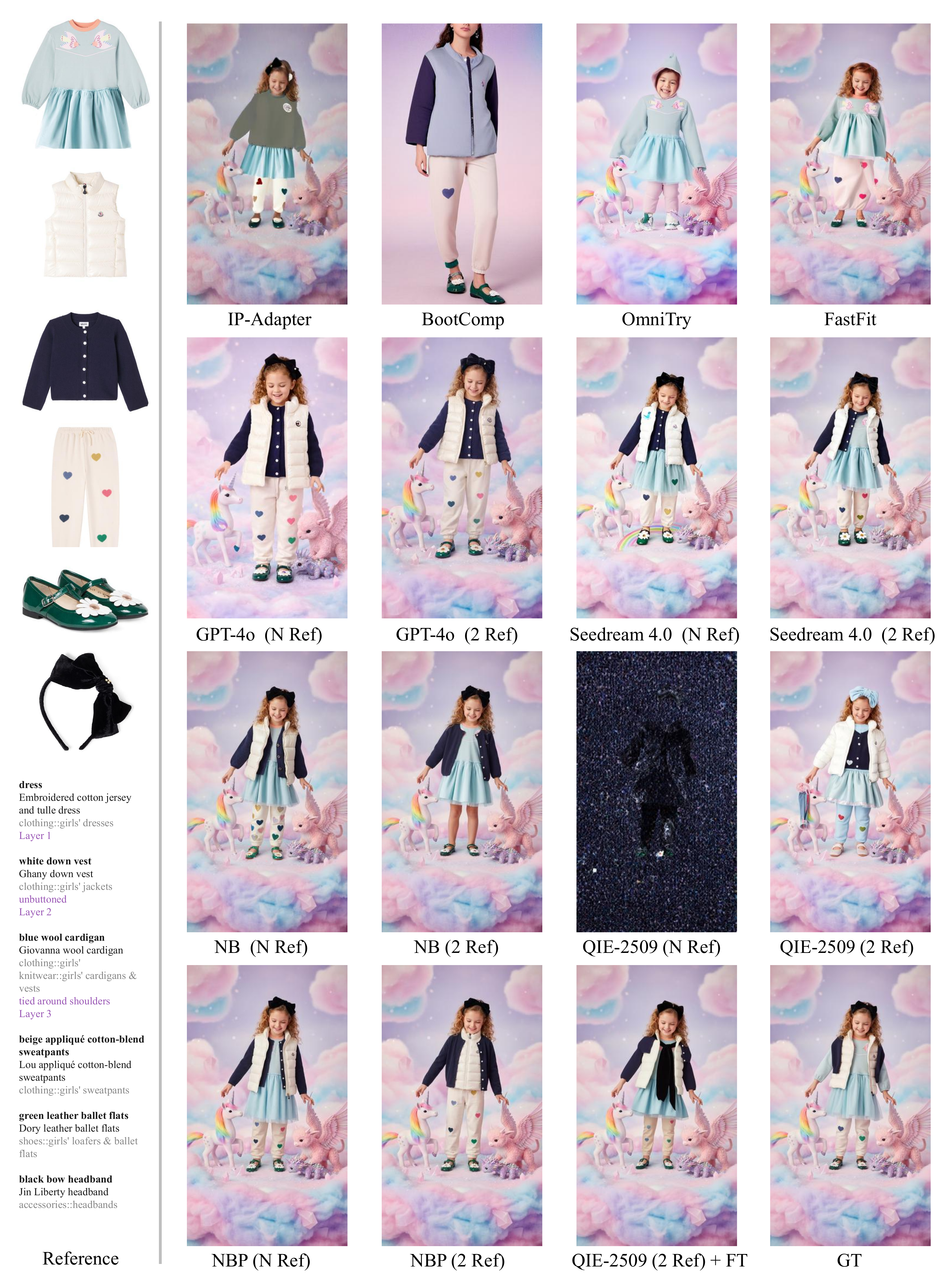}
    \vspace{-1.2em}
    \caption{More comparison results on Garments2Look test set (3/5).}
    \label{fig:supp-more-comparison-results-3}
\end{figure*}

\begin{figure*}[t]
    \centering
    \includegraphics[width=0.94\linewidth]{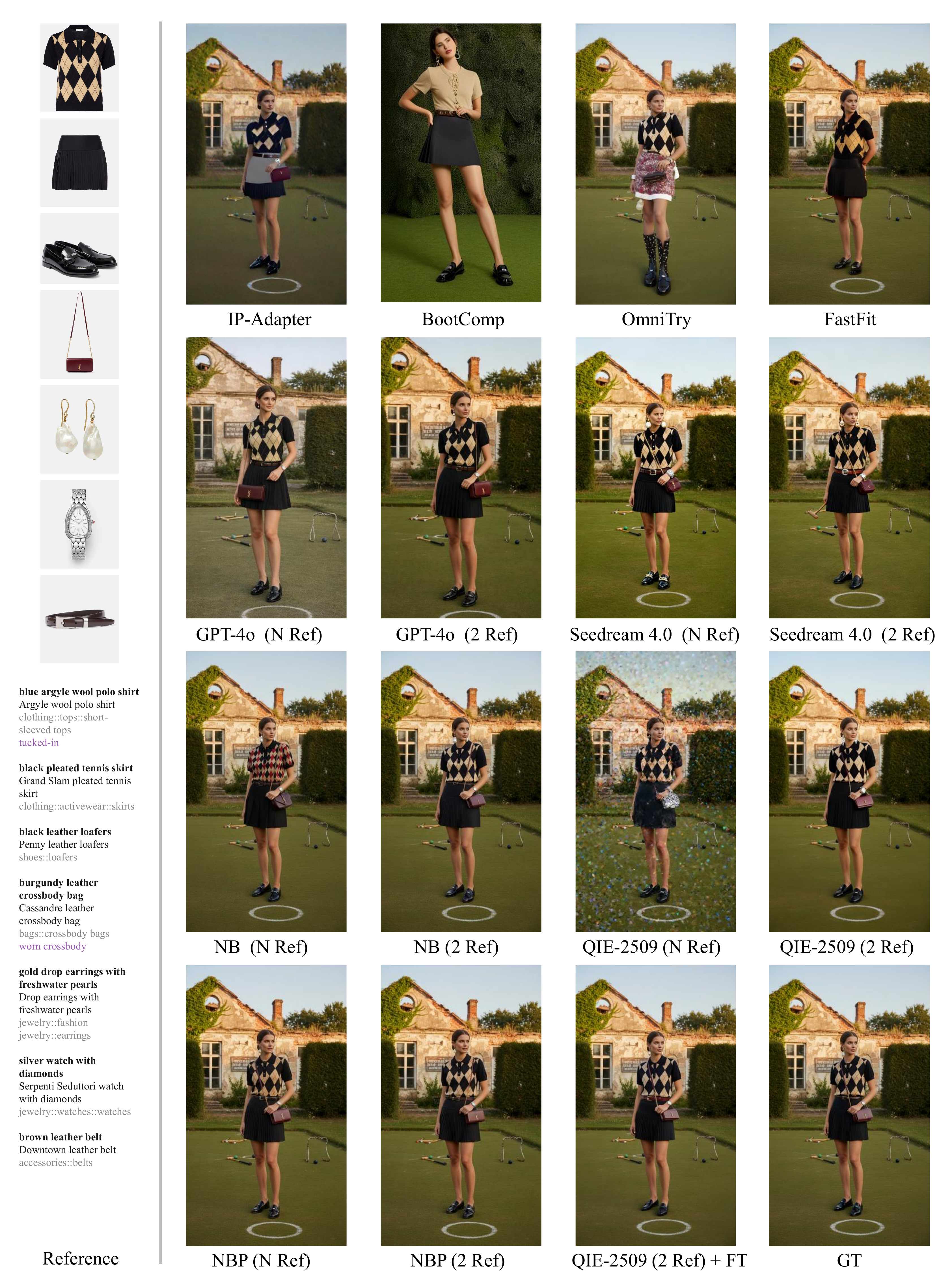}
    \vspace{-1.2em}
    \caption{More comparison results on Garments2Look test set (4/5).}
    \label{fig:supp-more-comparison-results-4}
\end{figure*}

\begin{figure*}[t]
    \centering
    \includegraphics[width=0.94\linewidth]{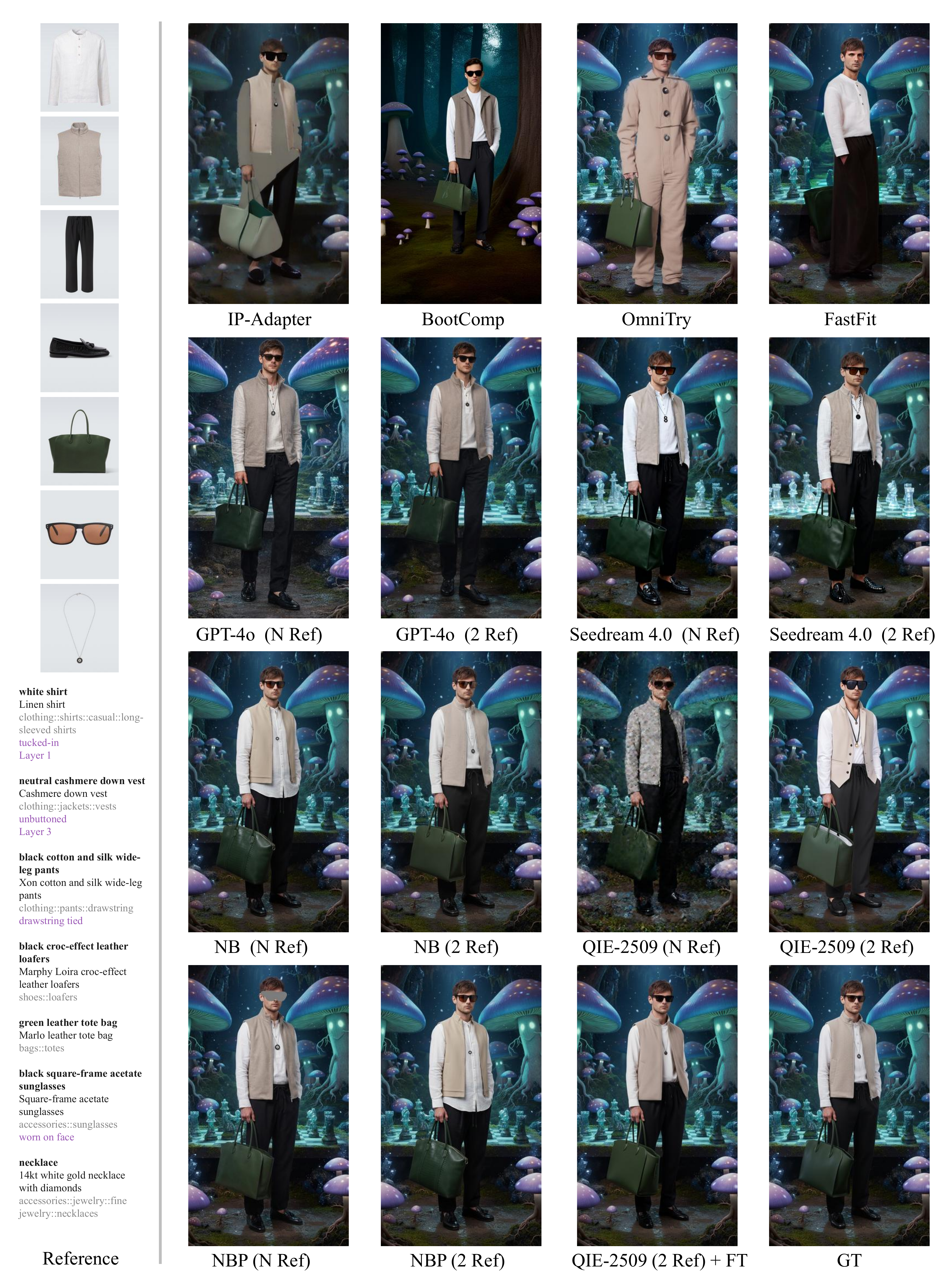}
    \vspace{-1.2em}
    \caption{More comparison results on Garments2Look test set (5/5).}
    \label{fig:supp-more-comparison-results-5}
\end{figure*}

In~\cref{fig:supp-more-example-data-1,fig:supp-more-example-data-2,fig:supp-more-example-data-3,fig:supp-more-example-data-4,fig:supp-more-example-data-5}, we present sample data from multiple datasets to clearly demonstrate the rich annotations provided by our dataset. 
In~\cref{fig:supp-more-comparison-results-1,fig:supp-more-comparison-results-2,fig:supp-more-comparison-results-3,fig:supp-more-comparison-results-4,fig:supp-more-comparison-results-5}, we show additional generated results from state-of-the-art VTON models, general-purpose image editing models, and our simply fine-tuning model.
These examples illustrate VTON performance of different models, the inherent difficulty of the new task, outfit-level VTON, and our dataset utility for this task.

\newpage
\begin{table*}[t]
\centering
\footnotesize
% \resizebox{\linewidth}{!}{
\begin{tabular}{p{0.4in}p{5.4in}}
\toprule
\multicolumn{2}{c}{\textbf{Style Guidance Example: Minimalist Style}} \\
\midrule

{\textbf{Template}} &

\# Outfit Guide for Minimalist style

This guide summarizes the core principles of Minimalist fashion, emphasizing intentionality, timelessness, and understated elegance through clean lines, neutral palettes, and high-quality materials.

\quad

\#\# I. Core Preferences and Prohibitions

$|$ Category $|$ Preference $|$ Prohibition  $|$

$|$ \textbf{Primary Colors} $|$ White, Black, Grey, Navy, Beige/Camel $|$ Bright neons, Loud primary colors (e.g., electric blue), Overly saturated jewel tones $|$

$|$ \textbf{Accents Colors} $|$ Olive Green, Muted Blush, Terracotta, Soft Khaki, Charcoal $|$ Fluorescent shades, Clashing bold hues, Iridescent fabrics $|$

$|$ \textbf{Patterns} $|$ Solid, Subtle Stripe (e.g., pinstripe, thin horizontal), Micro-check, Herringbone (subtle), Small geometric $|$ Large florals, Busy animal prints, Loud abstract, Polka dots, Ornate damask $|$

$|$ \textbf{Style/Fit} $|$ Clean lines, Tailored, Structured, Slightly oversized (but intentional), A-line, Straight-cut $|$ Bodycon, Overly baggy/shapeless, Excessive ruffles/frills, Unstructured asymmetry $|$

$|$ \textbf{Textures/Materials} $|$ Cotton, Linen, Wool (Cashmere, Merino), Silk, Leather (smooth), Tencel/Modal $|$ Cheap synthetics, Overly shiny/glossy, Coarse tweed, Heavily distressed denim $|$

$|$ \textbf{Accessories} $|$ Delicate gold/silver jewelry, Structured leather bag, Classic leather belt, Plain sunglasses, Simple watch, Silk scarf $|$ Chunky statement necklaces, Heavily embellished bags, Novelty accessories, Excessive layering of jewelry $|$

$|$ \textbf{Overall Vibe/Mood} $|$ Effortless Chic, Serene, Timeless, Polished, Intentional, Calm $|$ Trendy, Flashy, Ostentatious, Cluttered, Overly Casual/Sloppy $|$

---

\quad

\#\# II. Styling Pattern

This stylist's outfits emphasize intentional simplicity, high-quality foundational pieces, and a harmonious balance of form and function to create a polished and enduring aesthetic.

\quad

\#\#\# 1. Classic Outfit Examples (4 examples)

$|$ Style $|$ Structure (Item Components) $|$ Keywords $|$

$|$ \textbf{Effortless Everyday} $|$ White oversized button-down shirt + Tailored black wide-leg trousers + Minimalist leather loafers + Delicate gold hoop earrings + Structured leather tote bag $|$ Polished, Comfortable, Versatile $|$

$|$ \textbf{Sophisticated Professional} $|$ Camel wool blazer + Cream silk camisole + Navy straight-leg pants + Low-heeled leather pumps + Sleek minimalist watch $|$ Refined, Business-Ready, Streamlined $|$

$|$ \textbf{Elevated Casual} $|$ Grey cashmere crewneck sweater + Black A-line midi skirt + White minimalist sneakers + Small crossbody bag + Simple pendant necklace $|$ Relaxed, Chic, Understated $|$

$|$ \textbf{Modern Urbanit} $|$ Black mock-neck long-sleeve top + Light wash straight-leg jeans + Ankle boots + Black tailored trench coat + Oversized sunglasses $|$ Contemporary, Sleek, Urban $|$

\qquad

\#\#\# 2. Outfit Extended Rules Summary

- \textbf{Color Palette}: Dominated by neutrals (black, white, grey, beige, navy), often monochromatic or with subtle tonal variations. Accent colors are muted and used sparingly to maintain serenity and avoid visual clutter.

- \textbf{Layering Principle}: Strategic and purposeful, focusing on adding warmth, dimension, or textural interest without creating bulk. Layers should align cleanly and contribute to the overall streamlined silhouette, emphasizing clean lines and functionality.

- \textbf{Fit Requirements}: Predominantly clean, tailored, and comfortable. Items should skim the body without being restrictive, or be intentionally oversized for a relaxed yet structured feel. Avoid anything ill-fitting, overly tight, or excessively baggy.

- \textbf{Pattern Restriction}: Strict preference for solids. If patterns are used, they are extremely subtle, such as thin pinstripes, micro-checks, or very faint textures, ensuring they do not disrupt the minimalist aesthetic but rather add a subtle layer of interest.

- \textbf{Shoe/Bag Principle}: Footwear and bags are chosen for their classic design, high-quality materials, and functionality. Styles are clean, unembellished, and contribute to the overall polished and timeless look (e.g., leather loafers, simple pumps, structured totes).

- \textbf{Accessory Requirements}: Minimalist accessories are curated for their understated elegance and practical purpose. They are few in number, high in quality, and serve to complement rather than dominate the outfit. Delicate jewelry, classic watches, and plain sunglasses are preferred.

- \textbf{Overall Balance}: Achieved through careful consideration of proportion, texture, and color. Outfits strive for visual harmony, avoiding any single element that overwhelms or distracts from the clean lines and intentional simplicity, creating a sense of calm and effortlessness.

- \textbf{Scene Adaptability}: Outfits are designed to be highly versatile, easily transitioning between different settings (e.g., office, casual outings, evening events) with minimal adjustments, thanks to their timeless appeal and foundational nature.

    \\ \bottomrule

\end{tabular}
% }

\caption{The prompt template of minimalist style guidance.}
\label{tab:minimalist_style_guide_prompt}
\end{table*}

\newpage

\newpage

\newpage

\newpage

\newpage

\newpage

\newpage

\newpage

\newpage

\newpage

% {
%     \small
%     \bibliographystyle{ieeenat_fullname}
%     \bibliography{supp}
% }

\end{document}